%% file: main.tex
\documentclass[10pt,twocolumn,letterpaper]{article}

\def\hidechanges{}

\usepackage[pagenumbers]{cvpr}

\input{preamble}

\definecolor{cvprblue}{rgb}{0.21,0.49,0.74}

\usepackage[
    pagebackref,
    breaklinks,
    colorlinks,
    allcolors=cvprblue
]{hyperref}

\usepackage{booktabs}
\usepackage{float}
\usepackage{amsmath,amssymb}
\usepackage{pifont}
\usepackage{xspace}
\usepackage{algorithm}
\usepackage{algpseudocode}
\usepackage{listings}

\def\paperID{11202}
\def\confName{CVPR}
\def\confYear{2027}

\newcommand{\papertitle}{
Indi2Go: A 360$^\circ$ Vision Dataset \\for
Two-Wheeler-Centric Autonomous Driving
in Mixed Traffic
}

\title{\papertitle}

\author{
Aditya Akhouri$^{*}$ \quad
Monalisa Ray$^{*}$ \quad
Kumar J. Jyothish \quad
Subhankar Mishra\\[6pt]
National Institute of Science Education and Research (NISER), Bhubaneswar, India\\[4pt]
{\tt\small adityakhouri@niser.ac.in \quad raymonalisa395@niser.ac.in}\\[1pt]
{\tt\small swayansaasita@niser.ac.in \quad smishra@niser.ac.in}\\[3pt]}

\begin{document}

\maketitle

\makeatletter
\begingroup
\renewcommand\thefootnote{}%
\long\def\@makefntext#1{\noindent #1}%
\footnotetext{$^{*}$These authors contributed equally to this work.}%
\endgroup
\makeatother

\begin{abstract}
      \input{sec/0_abstract}

      \par\smallskip\noindent
      \href{https://huggingface.co/datasets/smlab-niser/indi2go}
      {\texttt{[smlab-niser/indi2go]}}
  \end{abstract}

\input{sec/1_intro}

\input{sec/2_related}

\input{sec/3_setup}

\input{sec/4_results}

\section{Conclusion}
\input{sec/5_conclusion}

\section*{LLM Usage Statement}

We used OpenAI’s ChatGPT (Plus) and Anthropic’s Claude (Pro) for limited assistance with coding and pipeline implementation. We also used ChatGPT to refine the wording of selected manuscript passages. The authors independently identified the research gap, conceived the study, designed the methodology, and collected the data. The authors independently verified all research findings and experimental results.

\section*{Acknowledgments}
This work was supported by La Fondation Dassault Syst\`emes (funding agency), Grant ID IN-2023-1-10.

{
\small
\bibliographystyle{ieeenat_fullname}
\bibliography{references}
}

\clearpage
\renewcommand{\topfraction}{0.95}\renewcommand{\textfraction}{0.05}\renewcommand{\floatpagefraction}{0.6}
\renewcommand{\dbltopfraction}{0.95}\renewcommand{\dblfloatpagefraction}{0.6}
\setcounter{topnumber}{4}\setcounter{dbltopnumber}{4}\setcounter{totalnumber}{8}
\setcounter{section}{0}\setcounter{table}{0}\setcounter{figure}{0}\setcounter{equation}{0}
\renewcommand{\thesection}{S\arabic{section}}
\renewcommand{\thetable}{S\arabic{table}}
\renewcommand{\thefigure}{S\arabic{figure}}
\renewcommand{\theequation}{S\arabic{equation}}
\renewcommand{\theHsection}{supp.\arabic{section}}
\renewcommand{\theHtable}{supp.\arabic{table}}
\renewcommand{\theHfigure}{supp.\arabic{figure}}
\renewcommand{\theHequation}{supp.\arabic{equation}}
\maketitlesupplementary

\input{sec/6_Supplement_tc}

\end{document}

%% file: preamble.tex
\usepackage{float}
\usepackage{tikz}
\usetikzlibrary{arrows.meta,calc}
\usepackage{float}
\usepackage{placeins}

\usepackage{microtype}

\ifdefined\pdfsuppressptexinfo\pdfsuppressptexinfo=-1\fi
\ifdefined\pdftrailerid\pdftrailerid{}\fi
\AtBeginDocument{\hypersetup{pdfauthor={},pdftitle={},pdfsubject={},pdfkeywords={},pdfcreator={},pdfproducer={}}}
\definecolor{chgcolor}{named}{RoyalBlue}
\newif\ifshowchanges
\ifdefined\hidechanges\showchangesfalse\else\showchangestrue\fi
\DeclareRobustCommand{\added}[1]{\ifshowchanges{\color{chgcolor}#1}\else{#1}\fi}

\definecolor{pendcolor}{RGB}{214,96,0}
\ifdefined\hidepending
\newcommand{\pending}[2]{\ignorespaces}
\newcommand{\undercheck}[2]{\ignorespaces}
\else
\newcommand{\pendingwho}[1]{\if\relax\detokenize{#1}\relax\else\ (#1)\fi}
\newcommand{\pending}[2]{{\color{pendcolor}\sffamily\footnotesize [PENDING: #1\pendingwho{#2}]}}
\newcommand{\undercheck}[2]{{\color{pendcolor}\sffamily\footnotesize [UNDER CHECK: #1\pendingwho{#2}]}}
\fi

%% file: sec/0_abstract.tex
A scooter observes traffic from a low viewpoint that tilts through turns, with nearby road users visible beside and behind the rider. We introduce \textbf{Indi2Go}, a dataset of Indian mixed traffic collected from an electric scooter equipped with six calibrated external RGB cameras covering the front, sides, and rear. From approximately nine hours of raw recordings, we curate 1,080 scenes from 69 independent recordings, comprising 59,657 timestamped six-camera keyframes and 1.43 million images including intermediate sweeps. The dataset pairs these observations with camera calibration, timestamped ego poses estimated from GPS, IMU, and visual--inertial odometry, and recording-disjoint training, validation, and test splits. On 1,736 test windows from 14 held-out recordings, a multilayer perceptron using two seconds of offline ego-pose history reduces three-second average displacement error from 3.684\,m for constant velocity to 3.327\,m. A pose-history classifier attains 0.643 macro-F1 for lateral-motion anticipation, compared with 0.394 for the constant-turn-rate velocity baseline. We benchmark 2D object detection on 200 scenes with human-verified bounding boxes across 38 classes. We also evaluate monocular depth consistency against a road-plane reference and zero-shot transfer of car-trained detection and planning models.

%% file: sec/1_intro.tex
\section{Introduction}
\label{sec:introduction}

\begin{figure}[t]
    \centering
\includegraphics[width=0.8\columnwidth]{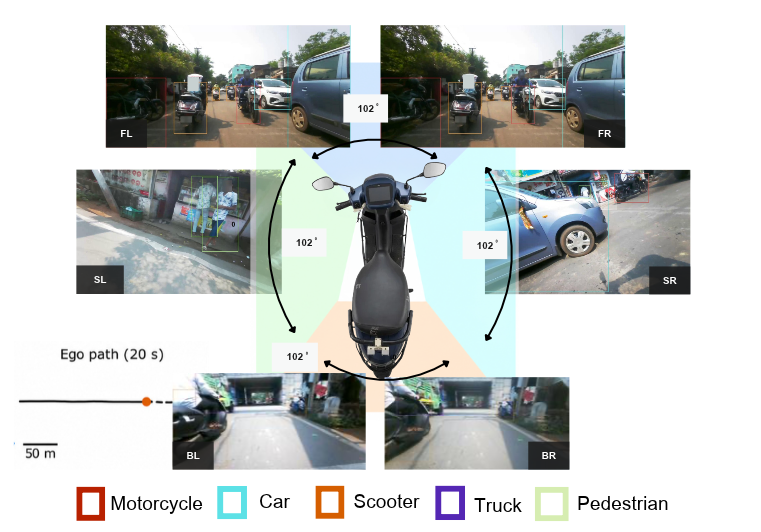}
    \caption{Indi2Go dataset example. The six cameras view , with 2D boxes coloured by class group; as well as the scooter's ego-pose trajectory over the 20\,s scene
  (solid: past, dashed: future, dot: this keyframe). Faces and plates are blurred.}
    \label{fig:scenarios}
\end{figure}

A scooter passing through a narrow gap may be close to vehicles on both sides while another road user approaches from behind. When the scooter leans into a turn, its cameras roll with it, changing the orientation of the road and the surrounding traffic in the image. For rider assistance, these are reasons to observe more than the road ahead. Calibrated front, side, and rear views, paired with the scooter's motion, let us study these interactions from the platform on which perception must operate.

Driving datasets such as KITTI, nuScenes, Argoverse, and Waymo have established benchmarks for perception and prediction from car-mounted sensors~\cite{geiger_are_2012,caesar_nuscenes_2020,chang_argoverse_2019,sun_scalability_2020}. WoodScape extends this setting to wide-angle surround imagery~\cite{yogamani2019woodscape}, while IDD and IDD-3D document heterogeneous road traffic in India~\cite{varma_idd_2019,dokania_idd-3d_2023}. Existing two-wheeler datasets address related questions. Real Track records motorcycle trajectories on a racetrack~\cite{vincke2024realtrack}, and IndiGo provides front and rear stereo, lateral depth sensing, and helmet imagery in Indian traffic~\cite{jyothish_multimodal_2025}. MOTOR and myEye2Wheeler study rider behaviour and gaze~\cite{paturkar_motor_2026,kumar_myeye2wheeler_2024}, while MotoVerse benchmarks rider-manoeuvre anticipation and the prediction of surrounding agents' trajectories~\cite{paturkar2026motoverse}. Indi2Go complements these resources by pairing six calibrated scooter-mounted RGB views and intermediate sweeps with a timestamped trajectory of the scooter itself.

We collect approximately nine hours of raw recordings from an electric scooter in Indian mixed traffic and curate 1,080 scenes from 69 independent recordings. The release contains 59,657 six-camera keyframes at a nominal rate of 3\,Hz and 1.43 million images including sweeps. Each camera stream carries a per-image timestamp on a common clock. We retain the original labelled keyframe images with corrected per-image timestamps and additionally provide nearest-time six-view groups selected from keyframes and sweeps, with per-view offsets and validity flags (Sec.~\ref{sec:supp_sync_audit}).Camera intrinsics and camera-to-scooter extrinsics accompany timestamped, recording-local ego poses estimated from GPS, IMU, and visual--inertial odometry. The train, validation, and test partitions keep all scenes from a recording together.

Our motion experiments ask how much the scooter's recent motion tells us about where it will go. On 1,736 windows from 14 held-out recordings, a multilayer perceptron using two seconds of offline pose history reduces three-second average displacement error from 3.684\,m for constant velocity to 3.327\,m. For lateral-motion anticipation, a pose-history classifier improves macro-F1 from 0.394 for constant-turn-rate velocity to 0.643.

Our contributions are fourfold.
\begin{itemize}
    \item We release \textbf{Indi2Go}, a calibrated six-camera surround-view dataset from an electric scooter in Indian mixed traffic, with timestamped ego poses, intermediate sweeps, and recording-disjoint evaluation splits. Held-out reprojection residuals characterize the calibration, and per-camera clock offsets with a per-sample validity flag describe the timing.

    \item We establish baselines for short-horizon scooter trajectory forecasting and lateral-motion anticipation, with a camera-coverage ablation in the supplement.

    \item We benchmark camera-level 2D object detection on 200 scenes with human-verified bounding boxes across 38 classes. We evaluate fine-tuned detectors, COCO-pretrained zero-shot checkpoints, and open-vocabulary detectors. For the remaining 880 scenes, we provide 2D pseudo-labels generated by the fine-tuned YOLO26-L checkpoint.

    \item We assess monocular depth consistency against a road-plane reference. We also examine zero-shot transfer from car-trained models by scoring projected 3D detections in the image plane and comparing UniAD's predicted ego trajectories with the recorded scooter motion~\cite{wang2021detr3d,wang2023streampetr,lin2023sparse4dv3,li2023bevdepth,hu_uniad_2023}.
\end{itemize}

%% file: sec/2_related.tex
\section{Related Work}
\label{sec:related_work}

\subsection{Driving Datasets and Surround-View Perception}

KITTI, nuScenes, Argoverse, and Waymo provide benchmarks for detection, tracking, and motion forecasting from car-mounted sensors~\cite{geiger_are_2012,caesar_nuscenes_2020,chang_argoverse_2019,sun_scalability_2020}. WoodScape uses four fisheye cameras to cover the surroundings of a car~\cite{yogamani2019woodscape}. IDD and IDD-3D document mixed Indian traffic and provide annotations for semantic and 3D perception~\cite{varma_idd_2019,dokania_idd-3d_2023}. Indi2Go captures mixed Indian traffic from a scooter, where the cameras sit low and roll with the platform through turns. This change in sensing geometry motivates our zero-shot transfer experiments with car-trained models.

\subsection{Two-Wheeler and Micromobility Datasets}

Real Track combines onboard imagery, inertial measurements, and positioning to reconstruct motorcycle trajectories on a racetrack~\cite{vincke2024realtrack}. IndiGo is closest in platform and traffic setting. It records Indian traffic from an electric scooter using front and rear stereo, lateral time-of-flight sensors, helmet imagery, and RTK positioning~\cite{jyothish_multimodal_2025}. DELTA uses an instrumented e-scooter to map sidewalks and pedestrian routes~\cite{zadegan2024delta}, while Oxford RobotCycle collects multimodal data for urban cycling safety and odometry~\cite{panagiotaki2025robotcycle}.

myEye2Wheeler studies rider gaze~\cite{kumar_myeye2wheeler_2024}, and RAAD supports single- and multi-view rider-intention prediction~\cite{gangisetty_icpr_2024}. MOTOR pairs front, rear, and helmet video with gaze and telemetry to study rider behaviour~\cite{paturkar_motor_2026}. MotoVerse benchmarks rider-manoeuvre anticipation and the future trajectories of surrounding road users~\cite{paturkar2026motoverse}. Table~\ref{tab:dataset_comparison} summarizes the sensing and annotation scope of selected datasets.

\begin{table*}[t]
\centering
\caption{Comparison of public two-wheeler datasets. Det.: external-object detection; Track.: tracking; Ego: GPS/IMU/telemetry ego-motion. DELTA is recorded from a stand-up e-scooter; its region is not stated in the paper. ``--'' indicates that the item is not a primary released annotation. Car datasets (KITTI, nuScenes, Waymo, IDD and others) are discussed in the text. $^{\ast}$Raw, timestamped iPhone depth; not validated as metric depth for the six cameras (Sec.~\ref{sec:sensor_configuration}).}
\label{tab:dataset_comparison}
\scriptsize
\setlength{\tabcolsep}{4pt}
\renewcommand{\arraystretch}{1.0}
\resizebox{\textwidth}{!}{%
\begin{tabular}{lccccccl}
\hline
Dataset & Year & Visual views & Depth / LiDAR & Det. & Track. & Ego & Region \\
\hline
Real Track \cite{vincke2024realtrack}
& 2024 & 360$^{\circ}$ handlebar camera & No & -- & -- & Yes & France (track) \\
myEye2Wheeler \cite{kumar_myeye2wheeler_2024}
& 2024 & Head-worn forward camera & No & -- & -- & -- & India \\
RAAD / ICPR-RIP \cite{gangisetty_icpr_2024}
& 2024 & Front + side-mirror cameras & No & -- & -- & -- & India \\
DELTA \cite{zadegan2024delta}
& 2024 & 4K + stereo cameras (e-scooter) & LiDAR & -- & -- & Yes & -- \\
IndiGo \cite{jyothish_multimodal_2025}
& 2025 & Stereo + lateral + helmet & Depth / PC & -- & -- & Yes & India \\
MOTOR \cite{paturkar_motor_2026}
& 2026 & Front + rear + helmet & No & -- & -- & Yes & India \\
\hline
\textbf{Indi2Go}
& \textbf{2026}
& \textbf{6 surround cameras}
& \textbf{Raw iPhone LiDAR$^{\ast}$}
& \textbf{2D (subset)}
& \textbf{--}
& \textbf{Yes}
& \textbf{India} \\
\hline
\end{tabular}%
}
\end{table*}

Indi2Go pairs six calibrated scooter-mounted RGB streams and timestamped sweeps with a recording-local ego trajectory. That trajectory provides the reference for forecasting the scooter's future displacement and classifying its realized lateral motion. The six views let us evaluate front, side, and rear visual coverage on the same prediction windows, with the pose history and targets held fixed.

%% file: sec/3_setup.tex
\section{Indi2Go Dataset}
\label{sec:setup}

Indi2Go pairs six calibrated external RGB views of Indian mixed traffic with timestamped scooter poses. Each recording has its own camera-to-iPhone clock offsets, local coordinate frame, and ego trajectory. The release links every keyframe and sweep image to its calibration and to the ego pose interpolated at its timestamp, with a validity flag.

\subsection{Sensor Setup}
\label{sec:sensor_configuration}

\begin{figure}[!tb]
    \centering
    \includegraphics[width=0.8\columnwidth]{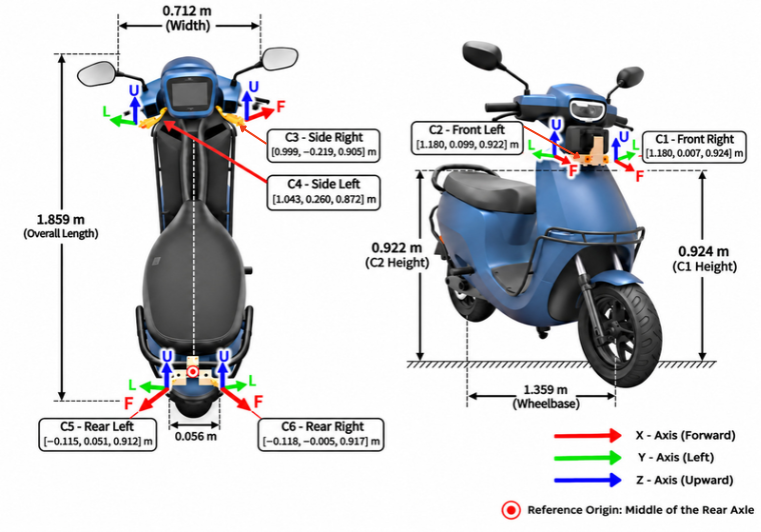}
    \caption{Surround-view sensor rig and coordinate-frame visualization, showing the six camera locations, sensor mounting position, and scooter dimensions with respect to the rear-axle reference origin. Legend: C:Camera}
    \label{fig:hardware_rig}
\end{figure}

We instrument an Ola S1 Pro electric scooter with six Arducam IMX708 USB cameras mounted approximately 0.9\,m above the road. The cameras are arranged in front, side, and rear pairs, with one camera on each side of the scooter in each pair (Fig.~\ref{fig:hardware_rig}). Each camera records an independent $1920\times1080$ H.264 stream at 30\,fps. A iPhone mounted at front of the bike jsut above the front cameras records CoreMotion IMU measurements at approximately 200\,Hz, GPS fixes at approximately 1\,Hz, and ARKit visual--inertial odometry (VIO). The iPhone supplies localization measurements. The released imagery comes from the six external cameras. The iPhone's LiDAR depth and confidence maps (approximately 60\,Hz) and its ARKit poses are released as a raw modality with their own iPhone-clock timestamps. iPhone RGB and intrinsics exist only for recordings from 6 June 2026 onward, and the iPhone-to-rig extrinsic calibration has not been validated for the 28--29 May recordings, so we do not use the LiDAR depth as metric ground truth for the external cameras.

The onboard recorder identifies the USB camera channels, starts the video encoders, and logs each camera's Unix start time. A companion iPhone application sends timestamped motion measurements. The recorder stores the six videos, timestamp tables, motion logs, and metadata together for each recording.

\subsection{Geometric Calibration}
\label{sec:coordinate_system}

We fitted a shared OpenCV rational intrinsic model to deduplicated
ChArUco observations pooled across the six camera channels. On the
fixed split of 139 fitting images and 35 held-out images, the model
achieved a training reprojection RMS of 0.869~px and a held-out
95th-percentile reprojection residual of 1.970~px (Tab.~S1).
Using surveyed planar targets, we estimated a separate rigid transform
from each camera to the scooter frame. The held-out 95th-percentile
extrinsic reprojection residuals ranged from 4.80 to 6.90~px across
the six cameras (Tab.~S2). The scooter frame is centred at the rear
axle midpoint, with \(+X\) forward, \(+Y\) left, and \(+Z\) up.
The released intrinsic matrices describe the images after
rectification, cropping, and resizing.

\subsection{GPS--IMU--VIO Ego-Pose Estimation}
\label{sec:ego_pose_estimation}

We estimate the scooter's six-degree-of-freedom trajectory offline for each recording in a local East--North--Up (ENU) frame. The first valid GPS fix defines the ENU origin. We screen fixes by reported horizontal accuracy, smooth the position reference with a second-order Savitzky--Golay filter over a 5\,s window~\cite{savitzky1964smoothing}, and interpolate it to the IMU timeline. We estimate the GPS--IMU time offset by matching changes in GPS course to integrated inertial yaw over valid turning intervals. The calibrated IMU-to-ego rotation maps gravity and angular rate into the scooter frame. Gravity-derived roll and pitch are smoothed over 0.5\,s. Gyroscope integration propagates yaw, and valid GPS course measurements correct the heading during motion with a 4\,s time constant.

Let $C_p$ denote the iPhone camera frame, $W_A$ the ARKit world frame, and $E$ the scooter frame. We convert the ARKit iPhone-camera pose to the scooter origin using the calibrated rigid transform $\mathbf T^{E}_{C_p}$:
\begin{equation}
    \mathbf T^{W_A}_{E}(t)
    =
    \mathbf T^{W_A}_{C_p}(t)
    \left(\mathbf T^{E}_{C_p}\right)^{-1}.
    \label{eq:dataset_arkit_ego}
\end{equation}
We form continuous VIO segments using tracking state, timestamp continuity, and motion consistency. Within each valid segment, a planar similarity fit registers horizontal VIO displacements to the GPS reference~\cite{umeyama1991least}. The rigid transform in Eq.~\eqref{eq:dataset_arkit_ego} changes the sensor reference point. The similarity fit registers the segment's horizontal trajectory.

At time step $k$, we propagate position using the registered VIO increment when available and the GPS-reference increment otherwise. We then apply a complementary correction towards the GPS-derived position reference $\mathbf p_k^{\mathrm{ref}}$:
\begin{equation}
\begin{aligned}
    \widehat{\mathbf p}_k
        &= \mathbf p_{k-1} + \Delta\mathbf p_k,\\
    \mathbf p_k
        &= \widehat{\mathbf p}_k
         + \left(1-e^{-\Delta t_k/\tau_p}\right)
           \left(\mathbf p_k^{\mathrm{ref}}-\widehat{\mathbf p}_k\right).
\end{aligned}
\label{eq:dataset_position_fusion}
\end{equation}
Here $\Delta t_k$ is the time between updates. The position correction uses $\tau_p=6$\,s within a valid VIO segment and $\tau_p=2$\,s otherwise. For attitude, relative rotations propagate the orientation, and spherical linear interpolation towards the GPS--IMU attitude reference applies the correction with time constants of 8\,s and 2\,s, respectively~\cite{shoemake1985animating}.

At each image timestamp on the iPhone clock, we linearly interpolate the fused position and use spherical linear interpolation for orientation. The external camera pose follows by composing this ego pose with the calibrated camera-to-ego transform. The per-scene motion files are derived from the same fused trajectory. Section~\ref{sec:supp_reference_fusion} gives the reference geometry and the heading update.

\subsection{Data Preprocessing and Organization}
\label{sec:temporal_alignment}
\label{sec:split}
\label{sec:dataset_partition}
\label{sec:data_organization}

\paragraph{Camera timing and frame selection.}
Each camera stream is timestamped. For every recording and camera, we estimate a camera-to-iPhone clock offset, which places the images on the clock of the IMU, GPS, ARKit, and fused ego pose. The original labelled keyframe images keep their grouping with corrected per-image timestamps; for multi-view use, the release additionally provides nearest-time six-view groups selected from keyframes and sweeps.Every accepted 3\,Hz keyframe contains one valid image from each of the six cameras. We select intermediate sweeps independently for each camera and retain their own timestamps. Each source video frame is extracted at most once. Per-camera offsets, the within-sample time spread, and a validity flag are released with every sample; Sec.~\ref{sec:supp_sync_audit} describes the timestamp audit and gives the spread statistics.

\paragraph{Image rectification and scene generation.}
We undistort each decoded image with the rational lens model, crop to the largest fully valid $16{:}9$ region, and resize the crop to $1920\times1080$ before JPEG export. We update the intrinsic matrix to account for the crop offset and resize factors. For each recording, we remove the first and last 5\,s of the common usable interval and divide the remaining data into nominal 20\,s scenes. A full scene can contain up to 60 six-camera keyframes. Intermediate images are linked to their associated keyframe while retaining their individual capture times.

\paragraph{Release structure and partitions.}
The \texttt{samples/} and \texttt{sweeps/} directories contain the rectified JPEGs. The JSON tables under \texttt{v1.0-trainval/} link recordings, scenes, and keyframes to individual images, calibrated sensors, and timestamped ego poses. The \texttt{ego\_motion/} directory contains per-scene pose, inertial, and reference metadata. Supplementary Fig.~\ref{fig:dataset_schema} shows the table relationships. The training, validation, and test partitions contain 756, 162, and 162 scenes from 38, 17, and 14 recordings, respectively. All scenes from a recording belong to the same partition.
\begin{table}[!tb]
\centering
\caption{Indi2Go acquisition and processed release. Scene duration and sample counts are nominal targets; totals correspond to exported data.}
\label{tab:dataset_statistics}
\footnotesize
\setlength{\tabcolsep}{4pt}
\begin{tabular}{lr}
\toprule
Property & Value \\
\midrule
Raw recording duration & $\sim9$\,h \\
Independent recordings & 69 \\
Scenes & 1,080 \\
Nominal scene duration & 20\,s \\
Six-camera keyframe rate & 3\,Hz \\
Nominal keyframes per full scene & 60 \\
External RGB cameras & 6 \\
Export image resolution & $1920\times1080$ \\
Six-camera keyframes & 59,657 \\
Keyframe images & 357,942 \\
Intermediate sweep images & 1,067,273 \\
Total camera images & 1,425,215 \\
Train / validation / test scenes & 756 / 162 / 162 \\
Train / validation / test recordings & 38 / 17 / 14 \\
\bottomrule
\end{tabular}
\end{table}

\begin{figure}[!tb]
    \centering
    \includegraphics[width=0.8\columnwidth]{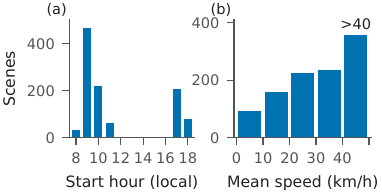}
    \caption{Distribution of the 1,080 scenes by (a) local recording start hour and (b) scene-average scooter speed.}
    \label{fig:scene_distribution}
\end{figure}

\section{Dataset Analysis}
\label{sec:data_analysis}
\label{sec:scene_diversity}
\label{sec:object_distribution}
\label{sec:ego_motion_analysis}

The 59,657 complete six-camera keyframes account for 357,942 images. A further 1,067,273 sweep images bring the release to 1,425,215 RGB images (Tab.~\ref{tab:dataset_statistics}). The keyframes provide regular six-view samples at a nominal 3\,Hz, while sweeps keep additional observations between keyframes.

\begin{figure}[!tb]
    \centering
    \includegraphics[width=0.86\columnwidth]{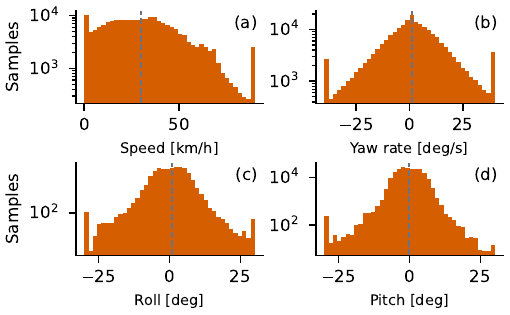}
    \caption{Estimated scooter speed, yaw rate, roll, and pitch across the dataset, derived from the fused ego-motion trajectory.}
    \label{fig:ego_motion_distribution}
\end{figure}

Figure~\ref{fig:scene_distribution} shows the distribution of scene start times and mean scooter speeds. The recordings span village streets, open roads, and dense mixed traffic. In these settings, side and rear views capture overtaking and following vehicles, while front views capture approaching and crossing traffic. Figure~\ref{fig:ego_motion_distribution} reports speed, yaw rate, roll, and pitch from the fused trajectory. The speed and yaw-rate distributions describe how the scooter moves, while roll and pitch describe how the camera rig tilts during those movements.

\begin{figure}[!tb]
    \centering
    \includegraphics[width=0.86\columnwidth]{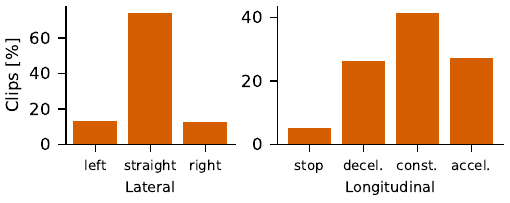}
    \caption{Dataset-wide distribution of trajectory-derived lateral and longitudinal movement over three seconds. The longitudinal distribution includes a separate stop category.}
    \label{fig:maneuver_distribution}
\end{figure}

Figure~\ref{fig:maneuver_distribution} summarizes realized lateral and longitudinal motion over three seconds. The descriptive longitudinal distribution includes a separate stop category. The supplementary camera-coverage experiment uses three speed-change classes defined in Sec.~\ref{sec:supp_motion_categories}.

%% file: sec/4_results.tex
\section{Benchmarks and Results}
\label{sec:evaluations}

\subsection{Evaluation Protocol}
\label{sec:experimental_protocol}

The in-domain experiments use the recording-disjoint split in Sec.~\ref{sec:dataset_partition}, with 756 training, 162 validation, and 162 test scenes. Each task uses the scenes and time windows that meet its input and reference requirements. The motion experiments use the offline GPS--IMU--VIO trajectory described in Sec.~\ref{sec:ego_pose_estimation}. Input states are timestamped at or before the forecast anchor, and the trajectory after the anchor supplies the targets. The pose histories inherit the offline smoothing and registration used to construct the released trajectories. The UniAD transfer experiment uses a separate protocol covering eligible anchors across the full release.

\paragraph{2D detection benchmark.}
The 2D detection benchmark uses 200 scenes from 21 recordings, split by recording into 125 training, 31 validation, and 44 test scenes. The benchmark vocabulary contains 38 classes, of which 29 have ground-truth boxes in the test split. We recover camera-local boxes from the six-view mosaics using the tile geometry and the correspondence to the released frames (Sec.~\ref{sec:supp_annotation_schema_notes}). We evaluate detection using COCO AP~\cite{lin2014coco}.

\subsection{Short-Horizon Ego-Motion Forecasting}
\label{sec:future_ego_motion}

We predict the scooter's planar motion over the next 3\,s from the preceding 2\,s of pose history. Each input contains seven states sampled at 3\,Hz, including the state at the forecast anchor. The target contains nine subsequent positions in the anchor-relative ego frame. Continuity and pose-quality checks yield 8,798 training, 1,980 validation, and 1,736 test windows. The test windows span 141 scenes from all 14 held-out recordings.

We compare constant velocity (CV), constant-turn-rate velocity (CTRV), and constant-turn-rate acceleration (CTRA) with three learned predictors: ridge regression, a gated recurrent unit (GRU), and a multilayer perceptron (MLP). All methods use the same test windows. We compute average displacement error (ADE) as the mean Euclidean position error up to the evaluation horizon and final displacement error (FDE) as the error at that horizon. We report both metrics at 1, 2, and 3\,s. Section~\ref{sec:supp_window_protocol} defines the coordinate transformation and metric computation.

\begin{table}[!htbp]
\centering
\caption{Future ego-trajectory forecasting on 1,736 matched test windows from 141 scenes and 14 held-out recordings. Errors in metres; lower is better.}
\label{tab:future_trajectory}
\footnotesize
\setlength{\tabcolsep}{4pt}
\resizebox{\columnwidth}{!}{%
\begin{tabular}{lrrrrrr}
\toprule
Method & ADE@1s & FDE@1s & ADE@2s & FDE@2s & ADE@3s & FDE@3s \\
\midrule
CTRA & 1.464 & 2.350 & 3.082 & 5.989 & 5.090 & 10.754 \\
CTRV & 1.413 & 2.240 & 2.832 & 5.309 & 4.463 & 8.993 \\
CV & 1.337 & 2.075 & 2.480 & 4.391 & 3.684 & 6.959 \\
Ridge (pose) & 1.340 & 2.143 & 2.594 & 4.707 & 3.888 & 7.380 \\
GRU (pose) & 1.320 & 2.046 & 2.430 & 4.306 & 3.587 & 6.714 \\
\textbf{MLP (pose)} & \textbf{1.187} & \textbf{1.860} & \textbf{2.236} & \textbf{4.024} & \textbf{3.327} & \textbf{6.267} \\
\bottomrule
\end{tabular}}
\end{table}

The MLP gives the lowest ADE and FDE at all three horizons (Tab.~\ref{tab:future_trajectory}). At 3\,s, it reduces ADE from 3.684\,m for CV to 3.327\,m, an improvement of 9.7\%, and reduces FDE from 6.959\,m to 6.267\,m. The MLP improves 3\,s ADE on 10 of the 14 test recordings. A paired bootstrap that resamples whole recordings gives a mean improvement of 0.357\,m with a 95\% confidence interval of $[0.047,0.658]$\,m. The supplementary breakdown shows that the gains are larger in turning and braking than in straight or mild motion (Sec.~\ref{sec:supp_window_protocol}).

\subsection{Future Lateral-Motion Anticipation}
\label{sec:future_lateral_motion}

We use the same 2\,s history to predict the scooter's realized lateral motion over the next 3\,s. Let $(\Delta x,\Delta y)$ be the terminal displacement in the anchor frame, with forward $x$ and leftward $y$. We assign a lateral event when $|\Delta y|\geq1.5$\,m, $\Delta x\geq2$\,m, and $|\operatorname{atan2}(\Delta y,\Delta x)|\geq8^\circ$. The sign of $\Delta y$ distinguishes leftward from rightward motion. All remaining windows are labelled minor change.

We compare a training-majority predictor, a CTRV classifier based on past motion, and a learned pose-history classifier on the same 1,736 test windows. The test set contains 295 leftward, 1,158 minor-change, and 283 rightward examples. We use macro-F1 and balanced accuracy so that each class contributes equally to the reported metrics.

\begin{table}[!htbp]
\centering
\caption{Future lateral-motion anticipation on 1,736 test windows (higher is better).}
\label{tab:future_lateral_motion}
\footnotesize
\setlength{\tabcolsep}{5pt}
\begin{tabular}{lrr}
\toprule
Method & Macro-F1 & Balanced acc. \\
\midrule
Training-majority class & 0.263 & 0.333 \\
CTRV (past motion) & 0.394 & 0.395 \\
\textbf{Pose-based classifier} & \textbf{0.643} & \textbf{0.661} \\
\bottomrule
\end{tabular}
\end{table}

The pose-history classifier reaches 0.643 macro-F1 and 0.661 balanced accuracy, compared with 0.394 and 0.395 for CTRV (Tab.~\ref{tab:future_lateral_motion}). Its validation macro-F1 is $0.639\pm0.008$ across three initialization seeds.

The supplement reports a camera-coverage ablation for both motion tasks. It also presents a NoPo4D~\cite{balice2026nopo4d} experiment on dynamic 4D reconstruction for temporal view interpolation.

\subsection{Monocular Road-Plane Depth Consistency}
\label{sec:depth_protocol}

We evaluate five pretrained monocular depth models without adaptation to Indi2Go: UniDepth V2~\cite{piccinelli2026unidepthv2}, Depth Pro~\cite{bochkovskiy2025depthpro}, MoGe-2~\cite{wang2025moge2}, Metric3D V2~\cite{hu2024metric3dv2}, and Depth Anything 3~\cite{lin2026depthanything3}. The reference depth is computed by intersecting calibrated image rays with an assumed gravity-aligned road plane using the released camera calibration and ego attitude. Mask2Former~\cite{cheng2022mask2former} supplies the road masks. We erode each mask by 15 pixels, sample with a six-pixel stride, and retain rays at least $3^\circ$ below the horizon with reference depths between 1 and 8\,m.

\begin{figure*}[h]
      \centering
      \includegraphics[width=\textwidth]{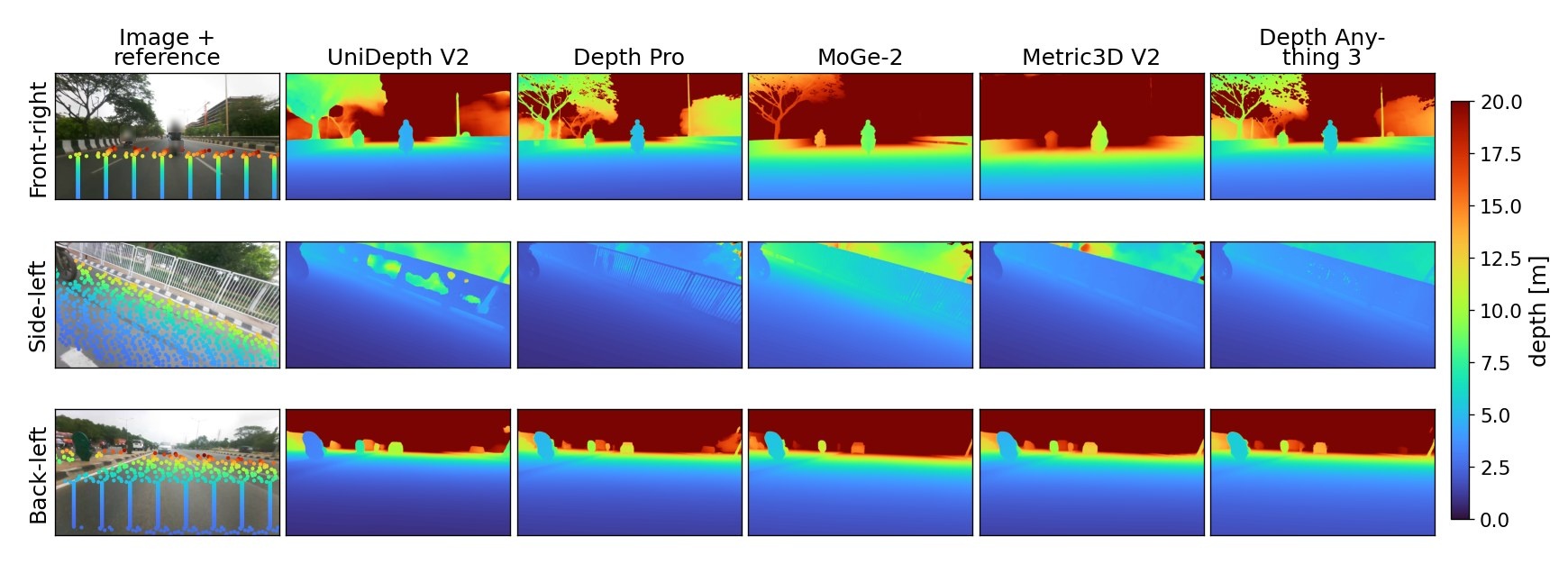}
      \caption{Qualitative monocular depth estimation results on
  representative driving scenes.}
      \label{fig:Depth_Estimation}
  \end{figure*}

{The evaluation covers approximately $1.4\times10^8$ road-reference points in 5,265 test images whose clock offset is confident and whose pose is valid (Sec.~\ref{sec:supp_sync_audit}).} We report agreement with the reference for the road-plane using absolute relative error (AbsRel), root mean squared error (RMSE), and threshold precision $\delta_1$. We evaluate predictions at their native scale (N) and after per-image median scaling to the reference (A). The latter measures agreement after accounting for per-image scale (Sec.~\ref{sec:supp_depth_reference}).

\begin{table}[!htbp]
\centering
\caption{Zero-shot monocular depth consistency with the road-plane reference on 5,265 images. A/N denote median-aligned/native scale. RMSE is in metres and $\delta_1$ is a percentage.}
\label{tab:depth_main}
\scriptsize
\setlength{\tabcolsep}{3.2pt}
\resizebox{\columnwidth}{!}{%
\begin{tabular}{lrrrrr}
\toprule
Method & A-AbsRel$\downarrow$ & A-RMSE$\downarrow$ & A-$\delta_1\uparrow$ & N-AbsRel$\downarrow$ & N-$\delta_1\uparrow$ \\
\midrule
UniDepth V2 & \textbf{0.069} & 0.541 & \textbf{94.8} & \textbf{0.166} & \textbf{68.9} \\                         Depth Pro & 0.072 & \textbf{0.448} & 94.5 & 0.210 & 62.0 \\                   MoGe-2 & 0.078 & 0.509 & 93.7 & 0.652 & 12.8 \\                                  Metric3D V2 & 0.090 & 1.323 & 92.9 & 0.722 & 29.6 \\                          Depth Anything 3 & 0.086 & 0.478 & 93.1 & 0.357 & 35.6 \\    
\bottomrule
\end{tabular}}
\end{table}

UniDepth V2 gives the lowest aligned AbsRel (0.069), the highest aligned $\delta_1$ (94.8\%), and the lowest native-scale AbsRel (0.166). Depth Pro gives the lowest aligned RMSE at 0.448\,m (Tab.~\ref{tab:depth_main}).     

\subsection{2D Road-User Detection}
\label{sec:det2d}

Indi2Go provides per-camera 2D bounding boxes across all 1,080 scenes. We use the 200 scenes with human-verified boxes for the benchmark. For the remaining 880 scenes, we ran the fine-tuned YOLO26-L checkpoint (\texttt{best.pt}) on the camera images and provide its predictions as pseudo-labels. These boxes can support semi-supervised detection~\cite{liu2021unbiasedteacher,xu2021softteacher}. TruckDrive also uses inferred 2D detections, combining them with manual 2D and 3D annotations in its refinement pipeline~\cite{ghilotti2026truckdrive}.

\paragraph{Benchmark protocol.}
We fine-tune YOLO26-M~\cite{redmon2016yolo,yolo26_ultralytics}, RF-DETR-M and RF-DETR-L~\cite{robinson2025rfdetr}, D-FINE-M~\cite{peng2025dfine}, and DEIMv2-X~\cite{huang2025deimv2} from COCO-pretrained checkpoints. The latter four configurations are based on DETR~\cite{carion2020detr}. Each detector processes one camera image at a time. For each of three seeds, we select the checkpoint by validation AP and score it against the human-verified test boxes. We also report the separate three-epoch YOLO26-L run used for pseudo-labeling. It uses a 960-pixel input and aspect-ratio-preserving padding.

We report COCO AP$_{50:95}$ over the 29 classes with test boxes. For the zero-shot comparison, we score both the COCO-pretrained and fine-tuned models on seven shared classes (cow, dog, bicycle, motorcycle, car, bus, and truck), denoted Z7 (Tab.~\ref{tab:main_v21}).

\par\medskip\noindent
\begin{minipage}{\columnwidth}
    \centering
    \footnotesize
    \setlength{\tabcolsep}{3.5pt}
    \makeatletter
    \def\@captype{table}
    \makeatother
    \caption{2D detection on Indi2Go. Scores are COCO AP points. Fine-tuned results above the separator are mean $\pm$ sample standard deviation over seeds 0--2. YOLO26-L is a single run.}
    \label{tab:main_v21}
    \begin{tabular*}{\linewidth}{@{\extracolsep{\fill}}lccc@{}}
        \toprule
        & \textbf{29 classes} & \multicolumn{2}{c}{\textbf{Z7}} \\
        \cmidrule(lr){3-4}
        \textbf{Detector} & \textbf{Fine-tuned} & \textbf{Zero-shot} & \textbf{Fine-tuned} \\
        \midrule
        YOLO26-M  & $17.33 \pm 0.55$ & $31.04$ & $42.98 \pm 1.41$ \\
        RF-DETR-M & $20.31 \pm 0.22$ & $34.04$ & \textbf{48.36} $\pm 0.29$ \\
        RF-DETR-L & $19.44 \pm 0.79$ & \textbf{36.14} & $48.00 \pm 0.44$ \\
        D-FINE-M  & $18.75 \pm 0.70$ & $27.94$ & $46.00 \pm 0.38$ \\
        DEIMv2-X  & \textbf{21.23} $\pm 0.73$ & $33.57$ & $47.50 \pm 0.34$ \\
        \midrule
        YOLO26-L  & $13.88$ & $31.63$ & $34.69$ \\
        \bottomrule
    \end{tabular*}
\end{minipage}
\par\medskip

\paragraph{Detection results.}
DEIMv2-X has the highest mean AP over the 29 test classes (21.23). On Z7, RF-DETR-L has the highest zero-shot AP (36.14), and RF-DETR-M has the highest mean fine-tuned AP (48.36). Fine-tuning improves Z7 AP by 11.86--18.06 points across the five three-seed configurations, with the largest gain for D-FINE-M. In the separate YOLO26-L run, Z7 AP rises from 31.63 to 34.69 after fine-tuning.

\paragraph{Open-vocabulary detection.}
We prompt OWLv2~\cite{minderer2023owlv2} and Grounding DINO~\cite{liu2024groundingdino} with the Indi2Go class names. OWLv2, Grounding DINO Base, and Grounding DINO Tiny score 14.41, 11.15, and 4.47 AP over the 29 test classes. Their Z7 scores are 29.53, 29.88, and 13.74 AP, respectively.

\paragraph{Training data scale.}
D-FINE-M scores $18.59 \pm 0.55$ AP using 75\% of the training scenes and $18.75 \pm 0.70$ AP using all training scenes. The difference between the means is smaller than the variation across seeds at either fraction.

\subsection{Zero-Shot 3D Detection Transfer}
\label{sec:det3d_protocol}

We evaluate the public car-trained checkpoints of DETR3D~\cite{wang2021detr3d}, StreamPETR~\cite{wang2023streampetr}, Sparse4D
  v3~\cite{lin2023sparse4dv3}, and BEVDepth~\cite{li2023bevdepth} on the scooter's six camera streams without fine-tuning. Inference covers all 8,906
  keyframes in the 162-scene test partition, using the original keyframe groups, whose six images are not simultaneous (Sec.~\ref{sec:supp_sync_audit}).
  Quantitative scoring uses 4,037 human-reviewed camera-local boxes from 280 keyframes across six test scenes. The six scenes are few, so differences
  of a few AP points are not significant.

We project the predicted 3D box corners into each rectified camera and use their image-space enclosing rectangles for matching (Sec.~\ref{sec:supp_projected_detection}). We report overall projected-box AP at 2D intersection-over-union thresholds of 0.50 and 0.10, recall at 0.50, and AP$_{50}$ for four mapped classes: car, motorized two-wheeler, heavy vehicle, and pedestrian. The scores quantify image-plane agreement on the annotated subset.
 \begin{figure*}[h]
      \centering
      \includegraphics[width=\textwidth]{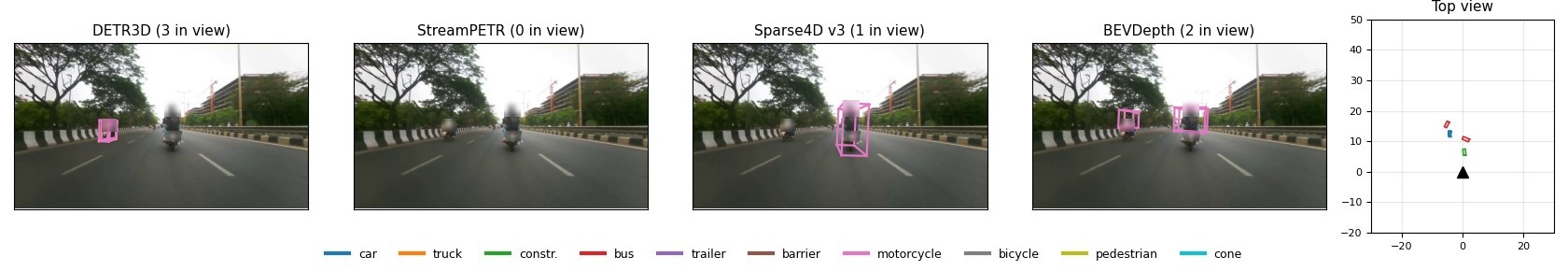}
      \caption{Zero-shot 3D detection transfer results on
  representative scooter-camera scenes, comparing predictions from
  DETR3D, StreamPETR, Sparse4D v3, and BEVDepth.}
      \label{fig:det3d_qualitative}
  \end{figure*}
  
\begin{table}[h]
\centering
\caption{{Zero-shot 3D detector transfer assessed by projected 2D boxes on 280 test keyframes (4,037 boxes), run on the original keyframe groups. AP/recall in \%. Upper: overall localization; lower: class AP50. mAP4: mean AP50 over the four classes. DETR3D uses R101-DCN; remaining methods use R50.}}
\label{tab:det3d_proj}
\footnotesize
\setlength{\tabcolsep}{4pt}
\begin{tabular}{lrrrr}
\toprule
Detector & AP$_{50}$ & AP$_{10}$ & Rec$_{50}$ & mAP$_4$ \\
\midrule
DETR3D & \textbf{14.8} & \textbf{37.4} & \textbf{26.1} & \textbf{9.7} \\
StreamPETR & 4.9 & 14.5 & 21.0 & 2.9 \\
Sparse4D v3 & 8.5 & 25.9 & 16.7 & 5.9 \\
BEVDepth & 0.7 & 11.3 & 6.6 & 1.2 \\
\bottomrule
\end{tabular}
\vspace{2pt}
\begin{tabular}{lrrrr}
\toprule
Detector & Car & Two-wheeler & Heavy & Pedestrian \\
\midrule
DETR3D & \textbf{24.4} & \textbf{7.9} & \textbf{4.6} & \textbf{1.6} \\
StreamPETR & 10.4 & 0.3 & 0.7 & 0.2 \\
Sparse4D v3 & 20.3 & 0.4 & 2.6 & 0.3 \\
BEVDepth & 3.9 & 0.0 & 1.0 & 0.0 \\
\bottomrule
\end{tabular}
\end{table}

DETR3D leads the evaluated checkpoints in projected AP$_{50}$ (14.8\%), AP$_{10}$ (37.4\%), recall at 0.50 (26.1\%), and four-class mAP$_{50}$ (9.7\%). Cars obtain the highest class AP$_{50}$ for every checkpoint. Even for DETR3D, AP$_{50}$ drops from 24.4\% for cars to 7.9\% for motorized two-wheelers, 4.6\% for heavy vehicles, and 1.6\% for pedestrians (Tab.~\ref{tab:det3d_proj}).

\subsection{Zero-Shot Ego-Trajectory Transfer}
\label{sec:zero_shot_protocol}

We apply the car-trained UniAD model~\cite{hu_uniad_2023} to scooter imagery without
  updating its weights. The inference adapter maps the six cameras to the expected input
  slots and supplies a fixed navigation command and the synthetic LiDAR reference transform
  required by the interface. The transform defines a coordinate frame for inference, and
  the recorded sensor inputs are RGB images. The model is run once on the original keyframe
  groups of all 59,657 samples across the 1,080 scenes. We compare its predictions with
  the scooter's reference trajectory, anchored at the corrected capture time of the
  front-left image, in an anchor-relative, motion-aligned frame at 0.5\,s intervals through
  3\,s (Sec.~\ref{sec:supp_uniad_transfer}).

  Of the 59,657 samples, 34,450 anchors have a valid corrected time, a valid pose, and a
  complete 3\,s future, and they enter the ADE, FDE, and terminal-heading evaluation. The
  stand-still reference, which predicts no motion, is evaluated on the same anchors. This
  full-release comparison uses 0.5\,s target intervals, while the in-domain forecasting
  benchmark uses its held-out 3\,Hz windows.

  \begin{figure}[h]
    \centering
    \includegraphics[width=\columnwidth]{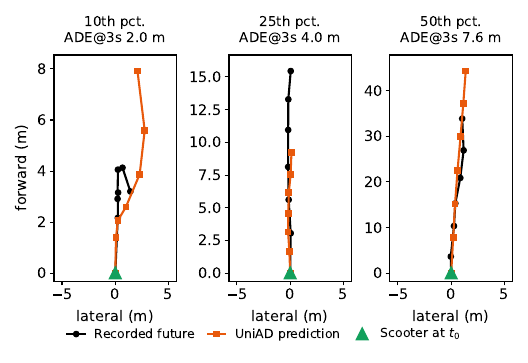}
    \caption{UniAD zero-shot ego-trajectory predictions against the recorded future at the 10th, 25th, and 50th percentiles of ADE@3s over 34,450 anchors. Each panel shows the predicted trajectory and the recorded future
path in the ego frame.}
    \label{fig:Zeroshot_Trajectory}
\end{figure}

  \begin{table}[!htbp]
  \centering
  \caption{Zero-shot UniAD ego-trajectory transfer (34,450 anchors with full 3\,s futures;
  errors in metres except heading in degrees).}
  \label{tab:zero_shot_main}
  \footnotesize
  \setlength{\tabcolsep}{3.2pt}
  \begin{tabular}{lrrrrr}
  \toprule
  Model & ADE@1s & ADE@2s & ADE@3s & FDE@3s & Head.@3s \\
  \midrule
  UniAD & 3.84 & 6.35 & 8.83 & 15.03 & 24.06 \\
  \bottomrule
  \end{tabular}
  \end{table}

  UniAD achieves ADEs of 3.84, 6.35, and 8.83\,m at 1, 2, and 3\,s, respectively. Its 3\,s
  FDE is 15.03\,m and its terminal heading error is $24.06^\circ$
  (Tab.~\ref{tab:zero_shot_main}). Its 3\,s ADE is lower than the 14.81\,m stand-still
  reference under the same transfer protocol (Sec.~\ref{sec:supp_uniad_transfer}). The six
  input images of each sample come from the original keyframe groups and are not
  simultaneous, so these values do not isolate model quality from input timing
  (Sec.~\ref{sec:limitations}).

\subsection{Limitations}
\label{sec:limitations}
The 2D detection benchmark uses human-verified boxes from 200 scenes. The detector-generated pseudo-labels for the remaining 880 scenes are provided separately, and the human review of the anonymized release is in progress. The six keyframe images of a sample are not simultaneous; the release provides corrected timestamps and nearest-time six-view groups. We report corrected-reference sensitivity analyses for 3D detectors and UniAD; time-consistent multi-view re-runs are identified separately in the supplement. The ego poses are session-local estimates from consumer GPS, IMU, and VIO without RTK ground truth, and the recordings contain no night-time data.

%% file: sec/5_conclusion.tex
We presented \textbf{Indi2Go}, a two-wheeler-centric dataset for perception in real-world Indian traffic. An electric scooter equipped with six rigidly mounted, calibrated RGB cameras captures forward, lateral, and rear views as the scooter moves through mixed traffic. The release pairs these views with timestamped ego poses estimated from GPS, IMU, and ARKit/VIO, and provides trajectory-derived motion labels and raw timestamped iPhone LiDAR data.

The 2D object detection benchmark uses 200 scenes with human-verified bounding boxes across 38 classes. Across five detector configurations evaluated with three seeds each, mean AP ranges from 17.33 to 21.23 over the 29 classes with test boxes, with DEIMv2-X achieving the highest mean. Fine-tuning improves AP by 11.86--18.06 points on the seven classes shared with COCO. For the remaining 880 scenes, the release provides pseudo-labels generated by the fine-tuned YOLO26-L checkpoint (\texttt{best.pt}).

In zero-shot transfer, the best car-trained 3D detector reaches 14.8\% projected AP$_{50}$, which measures image-plane agreement under our input adaptation. Indi2Go provides a basis for developing perception and motion prediction methods around the viewpoint, surround coverage, and dynamics of two-wheelers.

%% file: sec/6_Supplement_tc.tex
\section{Calibration and Ego-Pose Details}
\label{sec:supp_calibration_pose}

\subsection{Intrinsic Calibration}
\label{sec:supp_intrinsics}

All six external cameras use the same hardware and lens family. Calibration
uses a ChArUco target with \texttt{DICT\_5X5\_1000}, a $10\times7$ square
layout, square side $29.7$\,mm, and marker side $22.27$\,mm. Duplicate image
content is removed before model fitting and held-out evaluation.

For a raw camera, the intrinsic matrix is
\begin{equation}
\mathbf K=
\begin{bmatrix}
f_x & 0 & c_x\\
0 & f_y & c_y\\
0 & 0 & 1
\end{bmatrix},
\label{eq:supp_K}
\end{equation}
and the rational distortion vector is
\begin{equation}
\mathbf d=[k_1,k_2,p_1,p_2,k_3,k_4,k_5,k_6]^\mathsf T .
\label{eq:supp_distortion}
\end{equation}
For normalized coordinates $x=X/Z$, $y=Y/Z$, and $r^2=x^2+y^2$, the radial
factor is
\begin{equation}
L(r)=
\frac{1+k_1r^2+k_2r^4+k_3r^6}
     {1+k_4r^2+k_5r^4+k_6r^6}.
\label{eq:supp_rational}
\end{equation}

\begin{table}[H]
\centering
\caption{Held-out intrinsic model comparison on the deduplicated split.}
\label{tab:supp_intrinsic_comparison}
\footnotesize
\setlength{\tabcolsep}{4pt}
\begin{tabular}{lccc}
\toprule
Model & Train RMS [px] & Held-out P95 [px] & Selected \\
\midrule
OpenCV standard & 1.148 & 2.210 & no \\
OpenCV rational & 0.869 & \textbf{1.970} & yes \\
\bottomrule
\end{tabular}
\end{table}

The fisheye implementation did not return an estimated model because the
OpenCV call failed with incompatible array dimensions; it is therefore
reported as an implementation failure, not as a worse calibrated model.

\subsection{External-Camera Extrinsics}
\label{sec:supp_extrinsics}

Let $\mathbf T^E_{C_i}$ denote the rigid transform from camera $C_i$ to the
ego frame:
\begin{equation}
\widetilde{\mathbf p}^{E}
=
\mathbf T^E_{C_i}\widetilde{\mathbf p}^{C_i},
\qquad
\mathbf T^E_{C_i}=
\begin{bmatrix}
\mathbf R^E_{C_i} & \mathbf t^E_{C_i}\\
\mathbf 0^\mathsf T & 1
\end{bmatrix}.
\label{eq:supp_camera_extrinsic}
\end{equation}
The transforms are estimated from repeated observations of surveyed marker
mats on the plane $z_E=0$. Positive depth, expected viewing direction, and
cross-frame consistency reject mirrored planar solutions. The stored
quaternion, rather than the Euler summary below, is used during conversion.

\begin{table*}[!t]
\centering
\caption{Calibrated external-camera poses. Translation is the optical centre
in ego coordinates; P95 is the held-out reprojection residual.}
\label{tab:supp_camera_poses}
\footnotesize
\setlength{\tabcolsep}{4pt}
\begin{tabular}{lrrrrrrr}
\toprule
Camera & $t_x$ & $t_y$ & $t_z$ &
Roll [$^\circ$] & Pitch [$^\circ$] & Yaw [$^\circ$] & P95 [px] \\
\midrule
Front-left  & +1.180 & +0.099 & +0.922 & -91.2 & -1.4 & -90.5 & 4.80 \\
Front-right & {+1.180} & +0.007 & +0.924 & -91.1 & -2.6 & -90.8 & 5.09 \\
Side-left   & +1.043 & +0.260 & +0.872 & -110.4 & -14.4 & +16.6 & 6.90 \\
Side-right  & +0.999 & -0.219 & +0.905 & -110.1 & +18.6 & +164.8 & 6.48 \\
Back-left   & -0.115 & +0.051 & +0.912 & -104.2 & +0.7 & +91.2 & 5.50 \\
Back-right  & -0.118 & -0.005 & +0.917 & -104.2 & +2.3 & +90.2 & 4.96 \\
\bottomrule
\end{tabular}
\end{table*}

\begin{figure*}[!t]
    \centering
    \includegraphics[width=0.5\textwidth]{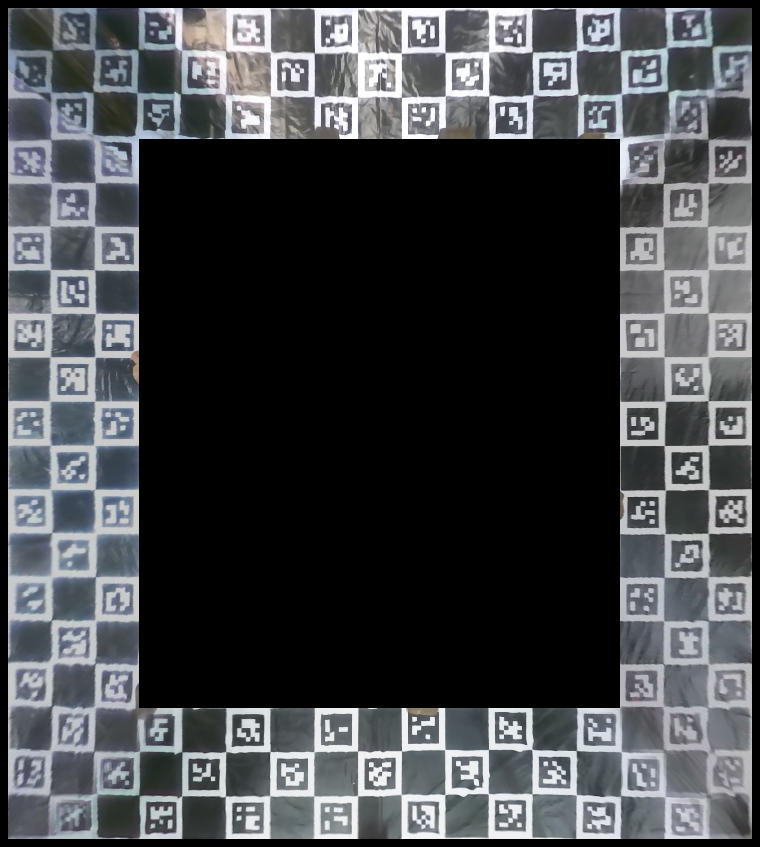}
    \caption{Metric-ring projection on the surveyed $z_E=0$ plane.
    }
    \label{fig:supp_ring_bev}
\end{figure*}

\begin{figure*}[!t]
    \centering
    \includegraphics[width=0.72\textwidth]{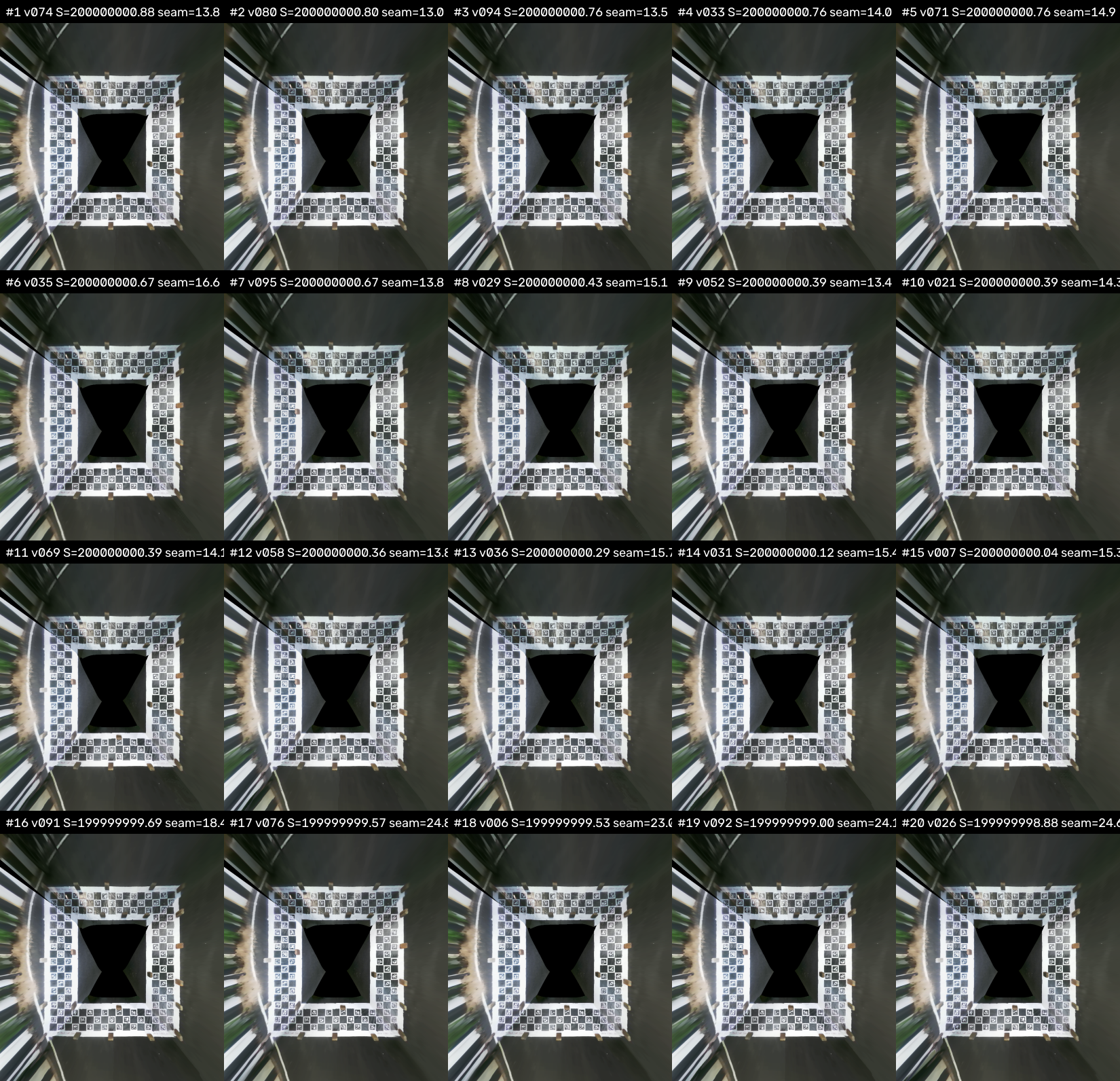}
    \caption{Road-surround visualization using the released image geometry,
    the six camera-to-ego transforms, and timestamp-specific ego poses.}
    \label{fig:supp_surround_bev}
\end{figure*}

\subsection{Camera Timing}
\label{sec:supp_timing}

Within a leaf, corrected camera time is represented by
\begin{equation}
\widehat t_i
=
t_i-\left[\delta_i+\gamma_i(t_i-t_0)\right],
\label{eq:supp_clock}
\end{equation}
where $\delta_i$ is the offset and $\gamma_i$ is the drift for camera $i${
relative to the front-left stream (a correction of a few milliseconds)}.
{Six-camera grouping is applied only at 3\,Hz keyframes.
Intermediate sweeps retain their own timestamps. The camera-to-iPhone clock
offset and the timestamp audit are described in Sec.~\ref{sec:supp_sync_audit}.}

\subsection{GPS--IMU--ARKit/VIO Ego Pose}
\label{sec:supp_ego_pose}

Let $C_p$ be the {iPhone} camera frame, $W_A$ the ARKit world frame,
$E$ the scooter ego frame, and $G$ the session-local ENU frame. ARKit
provides $\mathbf T^{W_A}_{C_p}(t)$, while calibration provides the rigid
device-camera-to-ego transform $\mathbf T^E_{C_p}$. The ARKit ego pose is
\begin{equation}
\mathbf T^{W_A}_{E}(t)
=
\mathbf T^{W_A}_{C_p}(t)
\left(\mathbf T^E_{C_p}\right)^{-1}.
\label{eq:supp_arkit_ego}
\end{equation}

A rigid alignment $\mathbf T^G_{W_A}$ maps the continuous ARKit segment to
the local ENU frame:
\begin{equation}
\mathbf T^G_E(t)
=
\mathbf T^G_{W_A}\,
\mathbf T^{W_A}_{C_p}(t)
\left(\mathbf T^E_{C_p}\right)^{-1}.
\label{eq:supp_enu_ego}
\end{equation}
When the GPS reference is displaced from the ego origin by
$\mathbf t^E_{\mathrm{GPS}}$, the ego translation is corrected as
\begin{equation}
\mathbf p^G_E(t)
=
\mathbf p^G_{\mathrm{GPS}}(t)
-
\mathbf R^G_E(t)\mathbf t^E_{\mathrm{GPS}}.
\label{eq:supp_gps_lever}
\end{equation}

The calibrated IMU-to-ego rotation maps measured gravity and angular rate:
\begin{equation}
\mathbf g^E(t)=\mathbf R^E_I\mathbf g^I(t),
\qquad
\boldsymbol\omega^E(t)=\mathbf R^E_I\boldsymbol\omega^I(t).
\label{eq:supp_imu_map}
\end{equation}
The final ego orientation combines the continuous ARKit/VIO heading with the
calibrated IMU tilt. At an image timestamp between pose samples
$t_a$ and $t_b$, translation is linearly interpolated and orientation uses
SLERP:
\begin{align}
\mathbf p^G_E(t)
&=(1-\alpha)\mathbf p^G_E(t_a)+\alpha\mathbf p^G_E(t_b),\\
\mathbf q^G_E(t)
&=\operatorname{SLERP}\!\left(
\mathbf q^G_E(t_a),\mathbf q^G_E(t_b),\alpha\right),\\
\alpha&=\frac{t-t_a}{t_b-t_a}.
\label{eq:supp_interpolation}
\end{align}
Finally, the global pose of external camera $C_i$ is
\begin{equation}
\mathbf T^G_{C_i}(t)
=
\mathbf T^G_E(t)\mathbf T^E_{C_i}.
\label{eq:supp_camera_world}
\end{equation}

The pose is session local. Ordinary {iPhone} GPS and ARKit/VIO do not establish
centimetre-level absolute world accuracy without independent RTK or surveyed
trajectory ground truth. Platform roll and pitch may be used for
roll-conditioned image evaluation, but they are not rider-body lean labels.

\begin{figure*}[!t]
    \centering
    \includegraphics[width=0.78\textwidth]{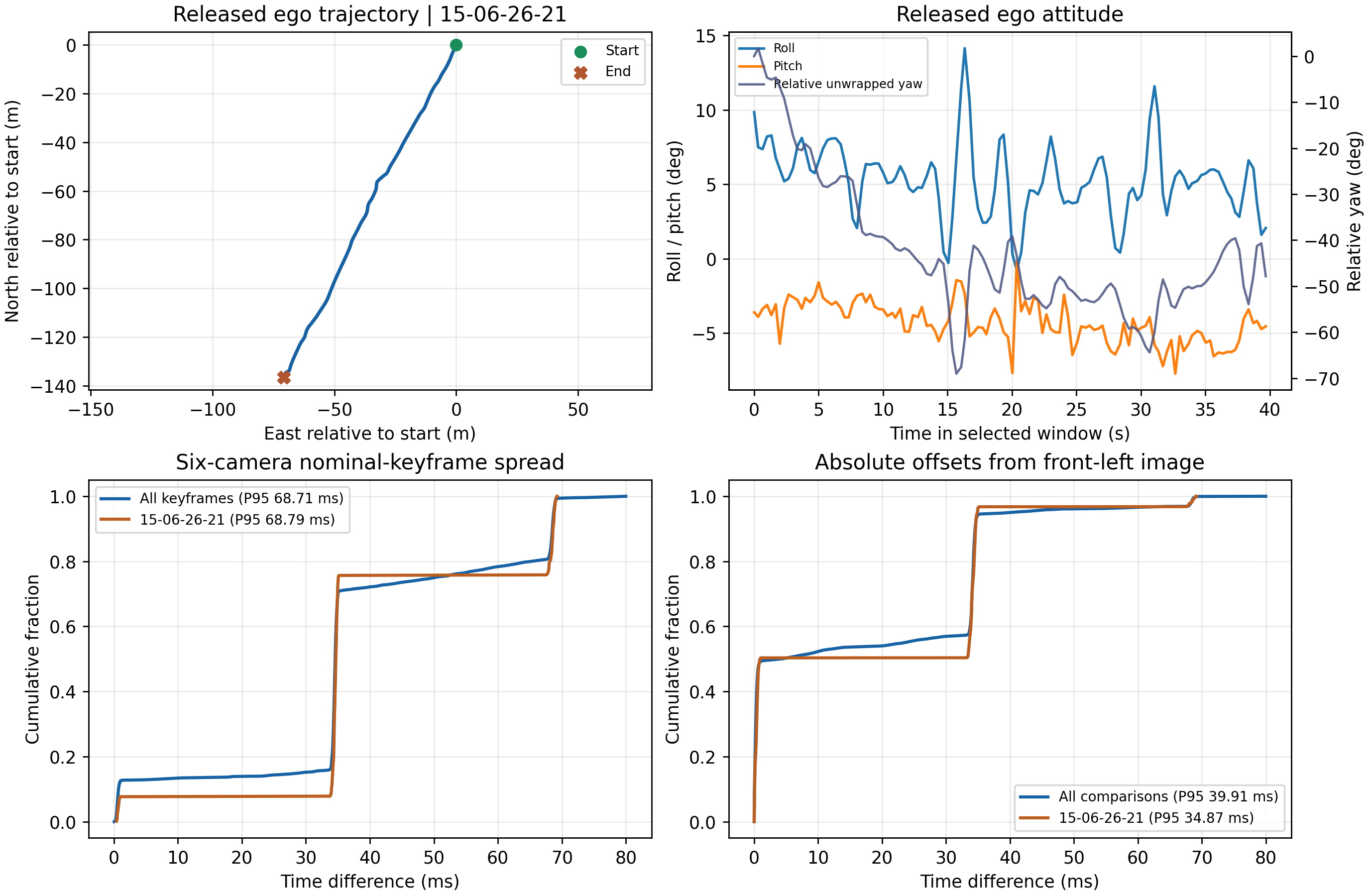}
    \caption{GPS, IMU, ARKit/VIO, fused session-local trajectory, and
    image-to-pose timing diagnostics.}
    \label{fig:supp_ego_graphs}
\end{figure*}

\subsection{Release Consistency}
\label{sec:supp_release_consistency}

For every committed scene, the converter verifies six external-camera
records per keyframe, valid media paths, valid calibrated-sensor references,
valid ego-pose references, and zero source-frame reuse. {iPhone RGB, iPhone
depth, and iPhone LiDAR are not registered as perception channels in these
tables; the raw iPhone depth, confidence, and ARKit poses are provided
separately with their own timestamps and without a validated extrinsic
calibration to the six cameras (Sec.~\ref{sec:sensor_configuration}).} These
checks verify database consistency; they do not substitute for independent
metric localization ground truth.

\section{Collection and Schema}
\label{sec:supp_collection}

\subsection{Recording Loop}
\begin{figure}[!t]
\centering
\includegraphics[width=\columnwidth]{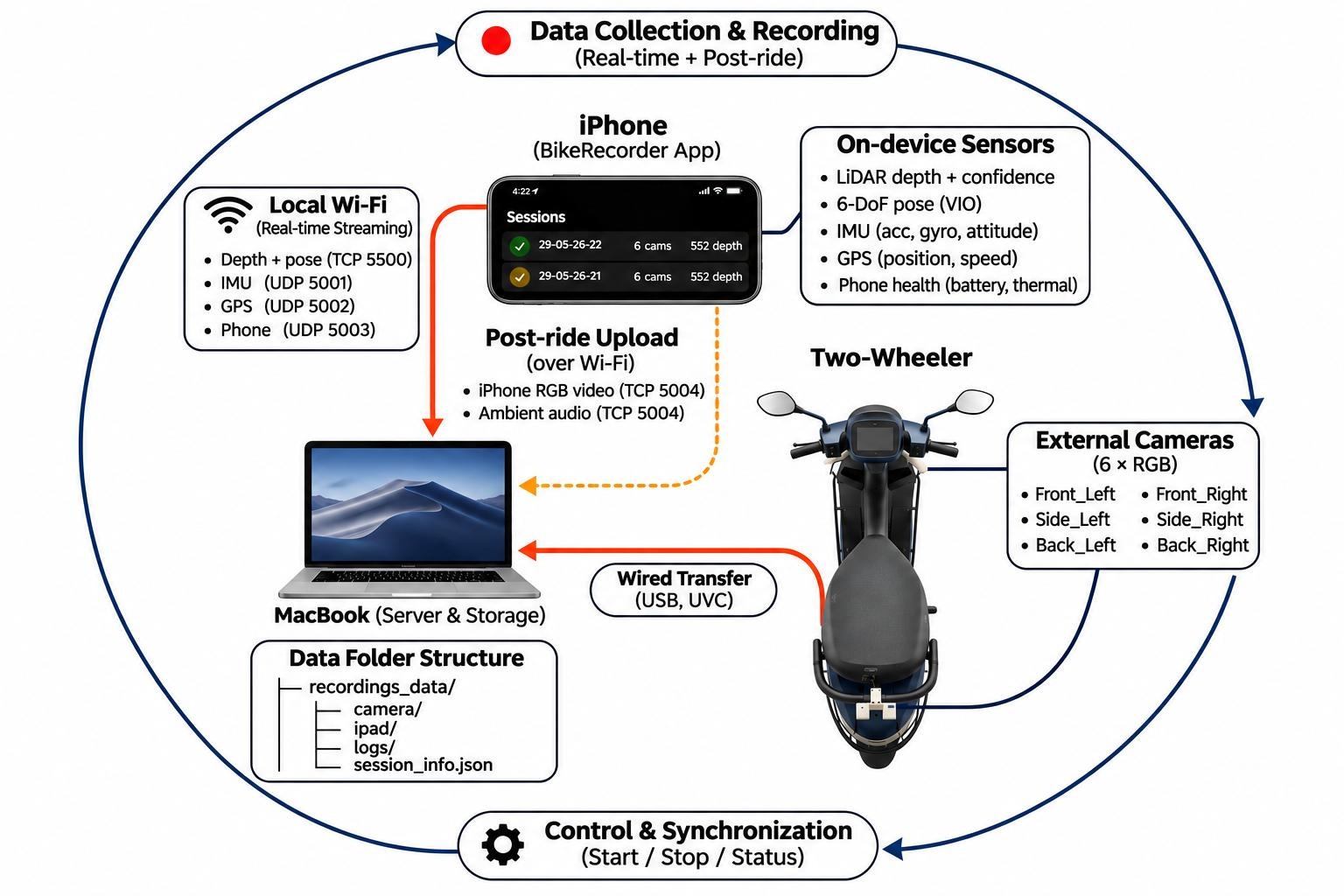}
\caption{Recording loop. The scooter carries the six cameras and the {iPhone}. The laptop runs {the recording script}, records the camera streams and sends commands to the iPhone, which streams its sensors back over WiFi and uploads its RGB video and audio after the ride. Session folders are then copied to a storage server.}
    \label{fig:collection_loop}
\end{figure}

The recording loop and the recorder are described in Sec.~\ref{sec:sensor_configuration} (Fig.~\ref{fig:collection_loop}).

\subsection{Dataset Schema}
\begin{figure*}[!t]
    \centering
    \includegraphics[width=0.85\textwidth]{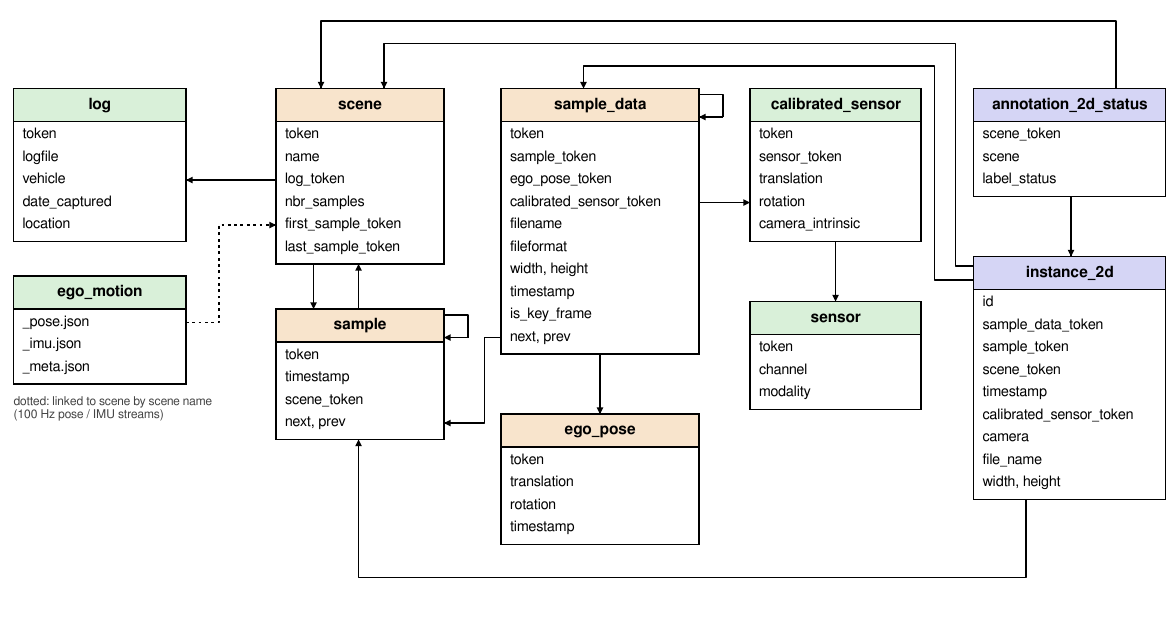}
    \caption{{Tables of the Indi2Go release (nuScenes-style) with their record counts. Italic fields are token references; arrows point to the referenced table. Annotation tables defined by the format are not populated in this release and are omitted.}}
    \label{fig:dataset_schema}
\end{figure*}

\paragraph{Core schema and 2D annotation extension.}
The relationships in Fig.~\ref{fig:dataset_schema} define image,
calibration, and ego-pose records. Camera-local 2D object annotations
are managed separately from the optional 3D-style annotation tables.
The latter are {defined by the format but not populated in this release, and are omitted from the illustration}; the
separate 2D labeling protocol is described in
Sec.~\ref{sec:supp_annotation_schema_notes}. The illustrated
\texttt{description} fields are optional schema fields and are not
required in the cleaned release JSON tables.

\subsection{Depth Qualitative Results}
\begin{figure*}[!t]
\centering
\includegraphics[width=0.85\textwidth]{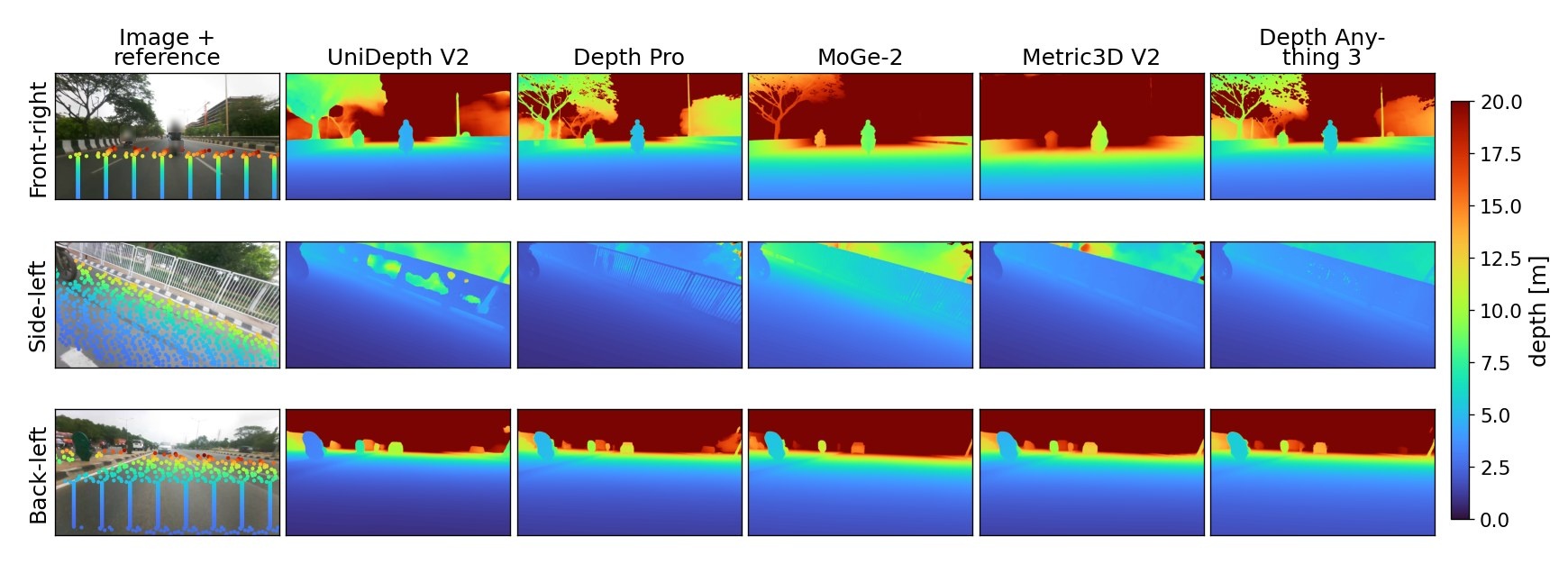}
\caption{Predicted depth (native scale, 0--20\,m) seen by the front-right, side-left and back-left cameras. The left column shows the
image with the road reference points (0--8\,m).}
\label{fig:depth_qual}
\end{figure*}

Figure~\ref{fig:depth_qual} shows predicted depth from three of the six cameras next to the road reference points.

\section{Camera Calibration: Model Selection and Quantitative Assessment}
\label{sec:supp_additional_calibration_evidence}

\subsection{Intrinsic Projection Models and Held-Out Residuals}

ChArUco observations employ ArUco fiducial-marker detection~\added{\cite{garridojurado2014aruco}}.

The calibration results in Table~\ref{tab:supp_intrinsic_comparison} correspond
to the final \emph{shared} intrinsic fit. Before freezing this fit, we also
investigated five projection-model families on the individual camera image
sets: Brown--Conrady rational, Kannala--Brandt equidistant, Mei unified
omnidirectional with radial--tangential distortion, Double Sphere, and the
Extended Unified Camera Model (EUCM). The exploratory comparison used held-out
corner-reprojection P95 as the primary ranking criterion, with held-out mean
error as a tie-breaker. The earlier per-camera results are reported in
Table~\ref{tab:supp_intrinsic_exploratory_models} to document the candidate
search; they are \emph{not} directly comparable with the final shared fit
because the calibration protocols and evaluation partitions differ.

\begin{table}[H]
\centering
\caption{Exploratory camera-model comparison from the earlier independent
per-camera diagnostic. The P95 range covers six camera channels. These
figures should not be compared directly with the shared-model errors in
Table~\ref{tab:supp_intrinsic_comparison}. All errors are in pixels.}
\label{tab:supp_intrinsic_exploratory_models}
\footnotesize
\setlength{\tabcolsep}{3pt}
\resizebox{\columnwidth}{!}{%
\begin{tabular}{lrrr}
\toprule
Model & Mean P95 & P95 range & Mean RMS \\
\midrule
Brown--Conrady rational & \textbf{1.56} & 1.37--1.81 & 0.83 \\
Kannala--Brandt equidistant & 1.76 & 1.49--2.09 & 0.97 \\
Mei unified omni-radtan & 3.19 & 2.73--3.54 & 1.74 \\
Double Sphere & 7.61 & 6.22--9.30 & 4.64 \\
Extended Unified Camera Model & 7.31 & 5.96--8.78 & 4.53 \\
\bottomrule
\end{tabular}}
\end{table}

The final comparison instead holds the deduplicated data split fixed across
the two admissible OpenCV variants. An audit of 264 source files identified
174 globally unique calibration images; 139 were assigned to fitting and
35 to held-out evaluation. The back-left and back-right folders contained
an exact duplicate image set, which was not counted twice. On this common
split, rational distortion decreased training RMS from 1.148 to 0.869\,px
and held-out P95 from 2.210 to 1.970\,px relative to the standard model,
corresponding to reductions of 24.3\% and 10.9\%, respectively. Selection was
thus supported by improved held-out reprojection consistency, not merely a
lower training residual. The shared lens model was adopted because the six
cameras use the same hardware and lens family 
fixed extrinsic pose. In contrast to the separately implemented
Kannala--Brandt diagnostic, the unsuccessful OpenCV fisheye call described
in Sec.~\ref{sec:supp_intrinsics} did not yield an estimated model and is
not assigned a quantitative rank.

\begin{table}[H]
\centering
\caption{Held-out corner-reprojection P95 of the accepted shared rational
intrinsic model, evaluated by source camera (pixels). This table describes
\emph{intrinsic} residuals; it is distinct from the extrinsic P95 values in
Table~\ref{tab:supp_camera_poses}.}
\label{tab:supp_intrinsic_channel_p95}
\footnotesize
\begin{tabular}{lr}
\toprule
Camera & P95 [px] \\
\midrule
Front-left & 1.970 \\
Front-right & 1.970 \\
Side-left & 1.522 \\
Side-right & 1.802 \\
Back-left & 2.485 \\
Back-right & 2.485 \\
\bottomrule
\end{tabular}
\end{table}

For a held-out image, calibration-board orientation and translation are
estimated using the fixed candidate intrinsics, and the detected ChArUco
corners are reprojected into that image. If $\mathbf u_{jk}$ and
$\widehat{\mathbf u}_{jk}$ denote a detected and reprojected corner,
respectively, its pixel residual is
\begin{equation}
 e_{jk}=\|\mathbf u_{jk}-\widehat{\mathbf u}_{jk}\|_2.
 \label{eq:supp_extra_corner_residual}
\end{equation}
The P95 statistic is the 95th percentile of these corner-wise residuals;
it is not a 95th-percentile uncertainty bound for arbitrary 3D scene points.
The held-out images are content-disjoint from fitting images but originate
from the same calibration corpus, so this result does not establish
independent cross-day calibration stability.

\begin{figure*}[!t]
\centering
\includegraphics[width=\textwidth]{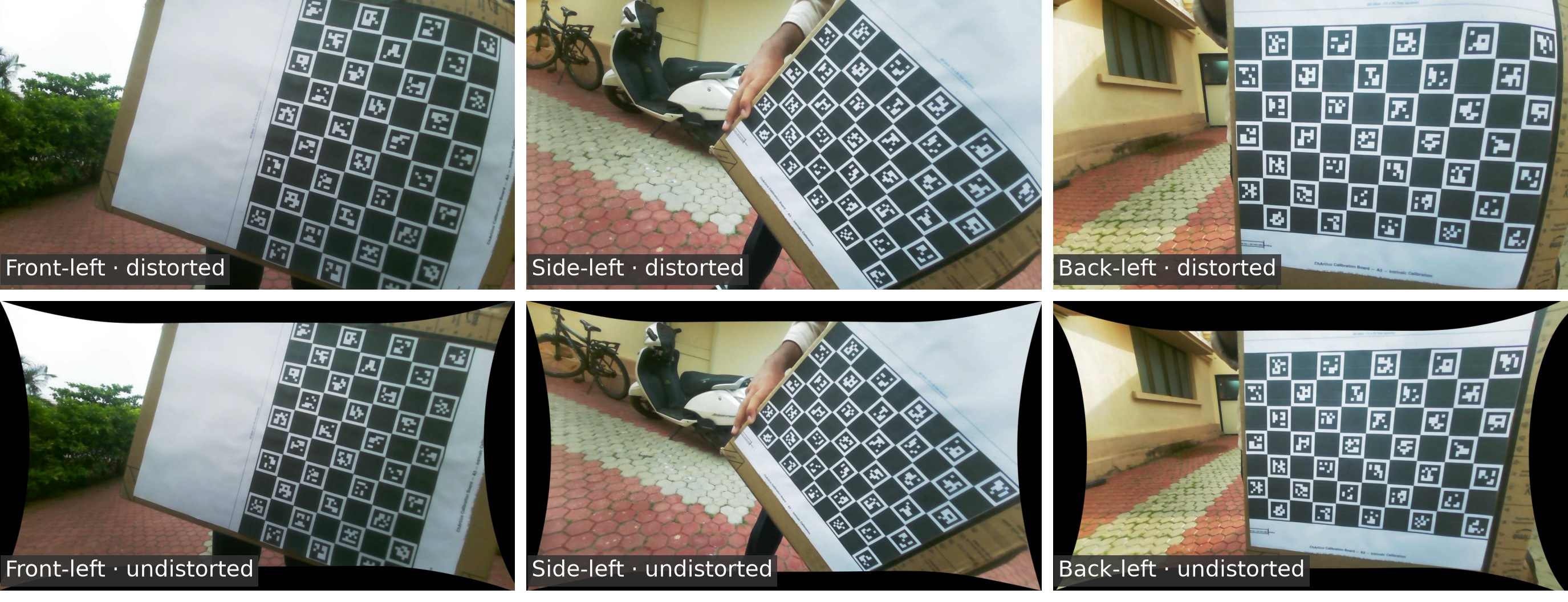}
\caption{ChArUco intrinsic-calibration examples for three cameras: original frame (top) and the same frame undistorted with the accepted rational model (bottom).These qualitative comparisons complement the held-out corner-reprojection statistics; they
are not an additional independent error measurement.}
\label{fig:supp_intrinsic_sixcam_panel}
\end{figure*}

\subsection{Camera-to-Ego Extrinsic Reprojection Analysis}

The optical-center locations and camera-to-ego rotations are reported in
Table~\ref{tab:supp_camera_poses}. They are fixed mounting parameters and do
not change with the scooter attitude. The six-camera reprojection analysis compares projections of known
metric-target points with their observed image positions in each of the
six external views. The accepted held-out
extrinsic P95 residuals span 4.80--6.90\,px across the six cameras. These residuals characterize the consistency of the fixed
camera-to-ego geometry with the metric target, independently of the
session-dependent GPS--IMU--VIO estimate.

\begin{figure*}[!t]
  \centering
  \includegraphics[width=\textwidth]{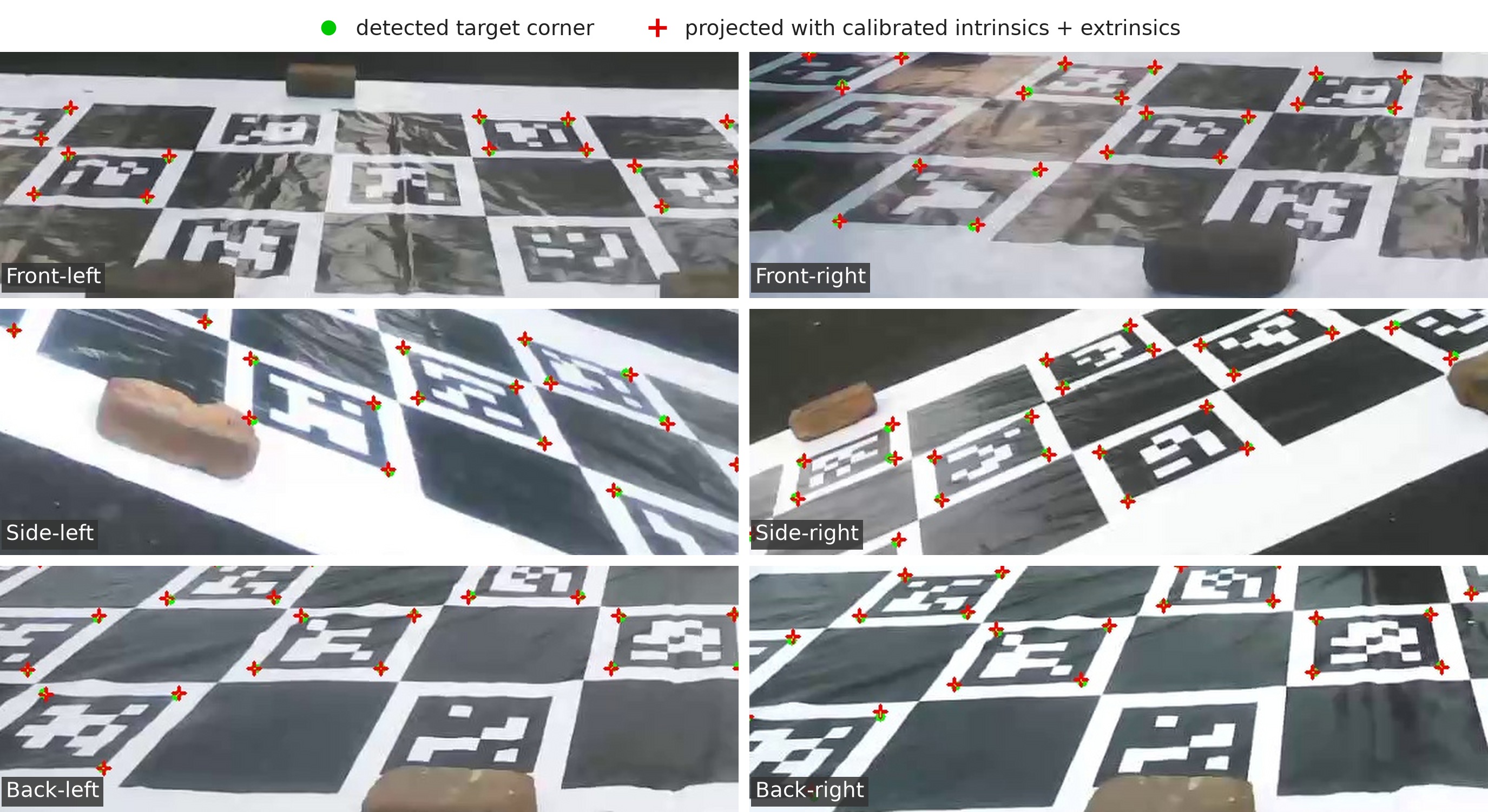}
  \caption{{Zoomed crops of the camera-to-ego reprojection check for the six cameras: detected target corners and their projections under the calibrated
  rig geometry.} The displayed correspondences
  supplement the held-out extrinsic P95 errors of
  Table~\ref{tab:supp_camera_poses}, rather than the intrinsic model-selection
  experiment in Table~\ref{tab:supp_intrinsic_comparison}.}
  \label{fig:supp_extrinsic_sixcam_panel}
  \end{figure*}

\section{Ego-Pose Fusion, Timestamp Quality, and Reference Geometry}
\label{sec:supp_additional_pose_validation}

\subsection{Reference-Point Geometry and Fusion}
\label{sec:supp_reference_fusion}

The scooter ego reference is located at the rear-axle midpoint, whereas the
{iPhone} measurements originate at the {iPhone}/GPS mount. The calibrated
lever arm expressed in the ego frame is
\begin{equation}
\begin{aligned}
\mathbf t^E_{\mathrm{GPS}}&=
[1.195,-0.285,1.036]^\mathsf T\,\mathrm m,\\
\|\mathbf t^E_{\mathrm{GPS}}\|_2&=1.607\,\mathrm m.
\end{aligned}
\label{eq:supp_numeric_gps_lever}
\end{equation}
Equation~\eqref{eq:supp_gps_lever} transports a position reference from the
measurement origin to the scooter origin. Camera extrinsics and the calibrated
{iPhone}-camera-to-ego transform are handled by
Eqs.~\eqref{eq:supp_camera_extrinsic} and~\eqref{eq:supp_arkit_ego}.
This ordering maintains a single ego-coordinate convention for the fused
position and the camera poses delivered in \texttt{ego\_pose.json}.

The GPS latitude, longitude, and height are first expressed in a
recording-local East--North--Up frame anchored at its initial valid fix:
\begin{equation}
\mathbf g^G(t)=\begin{bmatrix}
R\cos\varphi_0[\lambda(t)-\lambda_0]\\
R[\varphi(t)-\varphi_0]\\
h(t)-h_0
\end{bmatrix},\quad R=6{,}378{,}137\;\mathrm m,
\label{eq:supp_enu_reference}
\end{equation}
where latitude and longitude are in radians. A second-order five-second
Savitzky--Golay filter~\cite{savitzky1964smoothing} smooths the GPS reference, and its coordinates are
interpolated to the IMU timeline. The calibrated IMU-to-ego rotation maps
measured gravity and gyroscope rates according to
Eq.~\eqref{eq:supp_imu_map}. The GPS--IMU time shift is estimated from
agreement between GPS course changes and integrated inertial yaw rate over
valid turning intervals. Gravity-derived roll and pitch are smoothed over a
0.5\,s interval. Gyroscope integration supplies the relative yaw increment;
eligible GPS course measurements provide a heading reference during motion:
\begin{align}
\widetilde\psi_k&=\psi_{k-1}+\omega_{z,k-1}\Delta t_k,\\
\psi_k&=\widetilde\psi_k+
(1-e^{-\Delta t_k/\tau_\psi})\,
\operatorname{wrap}(c_k-\widetilde\psi_k),
\qquad \tau_\psi=4\,\mathrm s.
\label{eq:supp_yaw_reference}
\end{align}
Here $c_k$ is the course-derived ENU heading, and the correction applies
only when the course observation meets the speed and validity criteria.

The ARKit/VIO poses are transformed from the {iPhone} camera to the scooter ego
frame and organized into continuous segments according to tracking state,
pose-timestamp continuity, and motion consistency. Segment-wise horizontal
VIO displacement is aligned to the GPS reference using a robust planar
orthogonal similarity transform~\cite{umeyama1991least} with estimated scale. Valid segments provide relative
pose increments; GPS position and inertial attitude act as anchoring
references. For position, the complementary update is
\begin{align}
\widehat{\mathbf p}_k&=\mathbf p_{k-1}+\Delta\mathbf p_k,\\
\mathbf p_k&=\widehat{\mathbf p}_k+
\alpha_k(\mathbf g_k^G-\widehat{\mathbf p}_k),
\qquad \alpha_k=1-e^{-\Delta t_k/\tau_p},
\label{eq:supp_source_fusion}
\end{align}
where $\Delta\mathbf p_k$ uses registered VIO increments within a valid
segment and the GPS-reference increment otherwise. The corresponding
position correction time constants are $\tau_p=6$\,s during continuous VIO
and $2$\,s otherwise. For attitude, relative source rotations propagate
the orientation, and the GPS/IMU reference provides an SLERP correction
with $\tau_q=8$\,s during VIO and $2$\,s otherwise. Camera-time poses are
obtained with the interpolation in Eq.~\eqref{eq:supp_interpolation};
the six external camera poses follow Eq.~\eqref{eq:supp_camera_world}. Quaternion orientation interpolation uses spherical interpolation~\cite{shoemake1985animating}.
The same fused pose sequence supplies per-scene ego-motion outputs.

\subsection{Timestamp Audit and Correction}
\label{sec:supp_sync_audit}

\paragraph{Temporal preprocessing and quality assessment.}
The preprocessing pipeline aligns the camera streams relative to
the front-left camera, selects six-camera keyframes at 3\,Hz, and
retains intermediate sweeps at up to 12\,Hz. Each image is associated
with its individual timestamp, calibration, and interpolated ego pose.
For a six-camera sample $k$, the within-sample timestamp spread is
\begin{equation}
\Delta t_k=\max_{i=1,\ldots,6}t_{i,k}
          -\min_{i=1,\ldots,6}t_{i,k}.
\label{eq:supp_time_spread}
\end{equation}

Our release-wide quality assessment covers all 59{,}657 keyframes
and 1{,}425{,}215 camera images. All keyframes contain six camera
observations. The stored timestamp spread has a minimum of
0.058\,ms, a median of 34.571\,ms, and 95th and 99th percentiles
of 68.708\,ms and 69.067\,ms, respectively.

Across all six cameras, the image timestamps agree with their
corresponding filename timestamps and associated ego-pose timestamps,
with zero missing pose references, invalid translations, or invalid
rotations. All stored quaternions satisfy the unit-norm check.

\paragraph{Ego-pose quality assessment.}
The median within-keyframe ego-position spread across the six
camera timestamps is 0.283\,m, with 95th and 99th percentiles
of 0.981\,m and 1.500\,m, respectively. Independent trajectory
checks along each camera stream yield median speeds of
8.327--8.343\,m/s and median angular rates of
3.213--3.222$^\circ$/s, with no nonpositive temporal intervals.
The audit also identifies 6{,}402 intervals outside the nominal
sampling range and 5{,}472 camera-specific intervals with estimated
speeds above 25\,m/s. These intervals can include repeated observations
of the same underlying trajectory anomaly across cameras.
These checks characterize the consistency and limitations of the
released pose estimates rather than absolute localization accuracy.

\paragraph{Additional timestamp audit and correction.}
As part of release quality assurance, we additionally audited
camera timestamps against the image content. The converter of
the initial release extracted each scene's images with an OpenCV
frame-index seek. On variable-frame-rate videos, this seek can
introduce a constant frame displacement within a scene--camera
stream. An analytic model reproduces 115 of 115 replays, and
pixel matching against the raw videos agrees with it on all
5{,}877 confidently matched keyframes checked.

For the corrected experimental evaluations, each image is
re-stamped with the camera time of the raw frame that it actually
contains plus the measured camera-to-iPhone offset, estimated
per recording and camera by cross-correlating image motion with
the iPhone IMU. The ego pose is re-interpolated at the new time,
with validity flags for unsupported pose intervals. The 6.0\%
of images (85{,}773 of 1{,}425{,}215) whose offset estimates
have low confidence are flagged.

Under the corrected capture-time estimates, nearest-time grouping
reduces the median spread of the original labelled groups from
2.2\,s to 75\,ms. Of the 59{,}657 samples, 76\% lie within
100\,ms and 96\% within 200\,ms, with 55\% satisfying both
the 100\,ms spread criterion and the six-camera clock-confidence
requirement.

\paragraph{Effect on results.}
The 2D detection benchmark (Sec.~\ref{sec:det2d}) scores
single camera-local images against their own boxes, so its
results are unaffected. Results that combine several views or
associate images with ego poses are re-checked with the corrected
timestamps (Secs.~\ref{sec:supp_surround_ablation},
\ref{sec:supp_depth_reference},
\ref{sec:supp_projected_detection},
and~\ref{sec:supp_uniad_transfer}).
The additional timing-corrected evaluations are distinguished
from the original release metadata throughout the analysis.

\subsection{Motion-Window Construction}

The motion benchmarks use timestamped fused poses to construct histories
and future references. Each recording retains its own ENU origin, and
window eligibility is checked against the required sampling and temporal
continuity. Motion windows are computed inside a recording and inherit the
original recording-disjoint data split. The offline fused pose sequence is
the reference used for training and evaluation, whereas future states are
used to construct targets rather than direct model inputs.

\section{Supplementary Benchmark Protocols and Qualitative Material}
\label{sec:supp_additional_forecasting}

\subsection{Trajectory-Window Construction}
\label{sec:supp_window_protocol}

For the in-domain motion tasks in Secs.~\ref{sec:future_ego_motion}
and~\ref{sec:future_lateral_motion}, each eligible anchor at time $t_0$
uses the preceding two seconds of scooter-state history at 3\,Hz and
a future trajectory over three seconds. The anchor-relative planar
reference is formed by rotating each ENU displacement into the anchor's
ego frame,
\begin{equation}
\mathbf p^{(0)}(t)=
\mathbf P_{xy}\mathbf R^G_E(t_0)^\mathsf T
\left(\mathbf p^G_E(t)-\mathbf p^G_E(t_0)\right),
\label{eq:supp_anchor_relative}
\end{equation}
where $\mathbf P_{xy}=[\begin{smallmatrix}1&0&0\\0&1&0\end{smallmatrix}]$
selects the planar components. The same recording-local pose sequence
supplies past-state inputs and held-out future targets; future states are
never inputs. The recording-disjoint splits and the filtered windows
are those reported in the main paper. Displacement metrics are defined
in Sec.~\ref{sec:future_ego_motion}, and the full method comparison is
reported only in Tab.~\ref{tab:future_trajectory}.

\subsection{Motion-Category Construction and Longitudinal Examples}
\label{sec:supp_motion_categories}

\begin{figure*}[!t]
\centering
\includegraphics[width=0.94\textwidth]{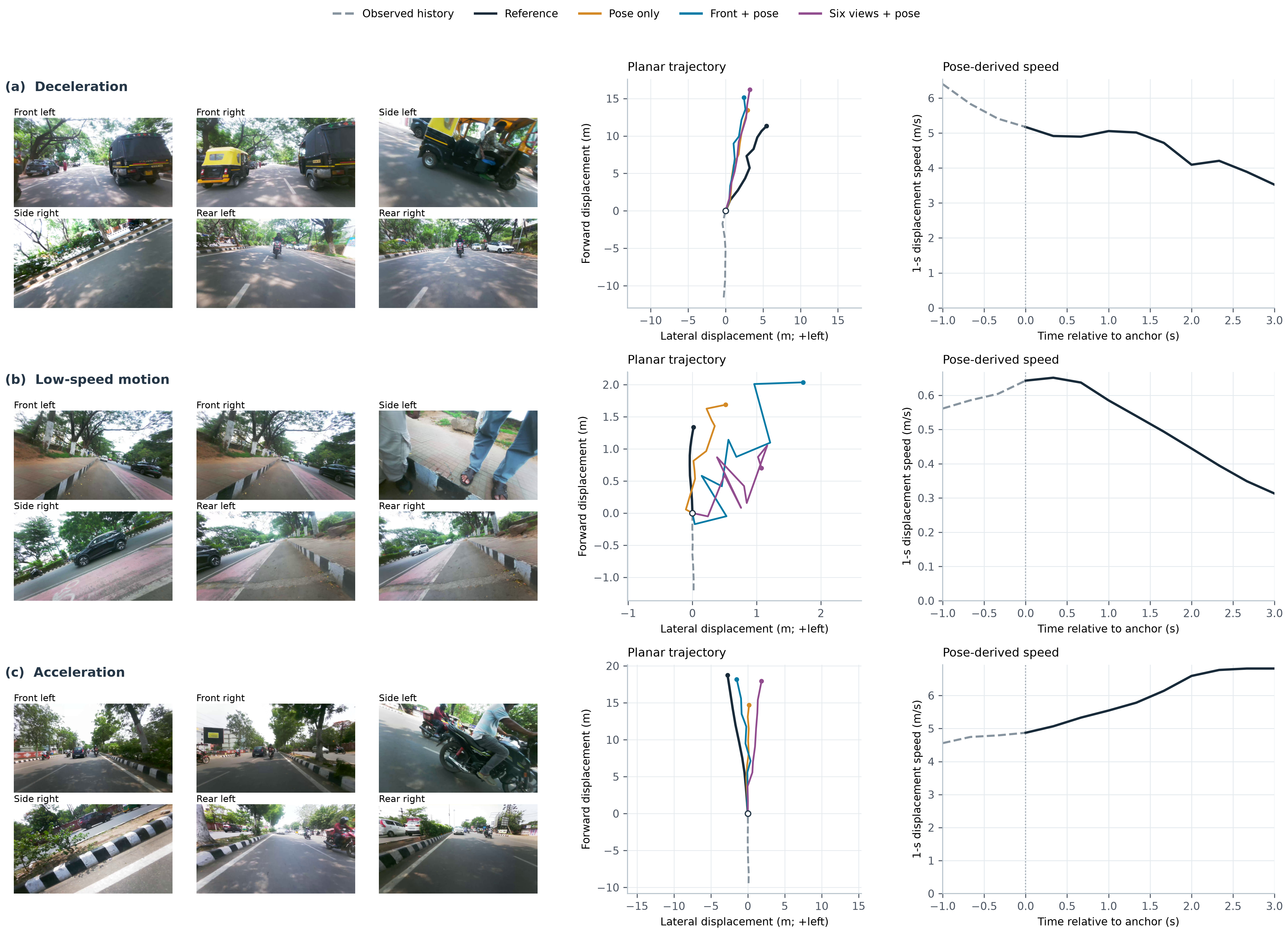}
\caption{Longitudinal scooter-motion examples from held-out recordings.
The camera views show surrounding traffic and the plots show the
past and future ego-motion used to construct the trajectory-derived
categories in Eq.~\eqref{eq:supp_longitudinal_rule}.}
\label{fig:supp_longitudinal_motion}
\end{figure*}

Lateral labels are assigned from the anchor-relative future motion by
the displacement and angular criteria in
Sec.~\ref{sec:future_lateral_motion}. For longitudinal categories,
let $\bar v_{h}$ and $\bar v_{f}$ denote the mean scooter speeds over
the final second of the observed history and of the future horizon,
respectively. The change
\begin{equation}
\Delta v=\bar v_f-\bar v_h
\label{eq:supp_longitudinal_rule}
\end{equation}
assigns \emph{deceleration} when $\Delta v<-1\,$m/s,
\emph{maintained speed} when $|\Delta v|\leq1\,$m/s, and
\emph{acceleration} when $\Delta v>1\,$m/s. The labels are constructed
from the observed trajectory for evaluation rather than inferred from
unobserved rider commands. Figure~\ref{fig:supp_longitudinal_motion}
illustrates the categories alongside their six-view visual context;
quantitative anticipation results {are in Tab.~\ref{tab:future_lateral_motion}
of the main paper and Tab.~\ref{tab:indi2go_surround_ablation}}.

\subsection{{Surround-View Context for Future Motion}}
\label{sec:surround_motion_ablation}
\label{sec:supp_surround_ablation}

We hold the 2\,s ego-state history and prediction targets fixed while varying camera coverage. The five configurations use pose history alone or combine it with front, front--side, front--rear, or all six camera views. For each included camera, we take the latest image at or before the forecast anchor and extract one descriptor with a frozen ImageNet-pretrained ResNet-18~\cite{he2016resnet}. A two-layer, four-head temporal Transformer~\cite{vaswani2017attention} combines the available descriptors with the pose history.

Each configuration predicts nine future planar positions and three-class lateral and longitudinal motion. The longitudinal label is based on the difference between mean speed in the final second of the forecast horizon and mean speed in the final second of the history. Changes below $-1$\,m/s indicate deceleration, changes above $+1$\,m/s indicate acceleration, and the remaining changes indicate maintained speed (Sec.~\ref{sec:supp_motion_categories}). We train each configuration with three seeds, select checkpoints on validation data, and evaluate on the same 1,736 held-out windows. The pose-only row is the reference for this multitask Transformer ablation.

\begin{table}[H]
\centering
\caption{Camera-coverage ablation: ADE@3s in metres ($\downarrow$), macro-F1 ($\uparrow$). Mean $\pm$ standard deviation across three seeds.}
\label{tab:indi2go_surround_ablation}
\scriptsize
\setlength{\tabcolsep}{2.6pt}
\begin{tabular}{lrrr}
\toprule
Input & ADE@3s & Lateral F1 & Longitudinal F1 \\
\midrule
\textbf{Pose history} & \textbf{$4.411\pm0.027$} & \textbf{$0.561\pm0.005$} & $0.539\pm0.023$ \\
Pose + front & $4.473\pm0.053$ & $0.511\pm0.016$ & $0.547\pm0.016$ \\
Pose + front/side & $4.619\pm0.047$ & $0.513\pm0.010$ & $0.548\pm0.005$ \\
Pose + front/rear & $4.509\pm0.009$ & $0.513\pm0.013$ & $0.544\pm0.015$ \\
Pose + all six & $4.564\pm0.107$ & $0.523\pm0.018$ & \textbf{$0.550\pm0.014$} \\
\bottomrule
\end{tabular}
\end{table}

Pose history alone gives the lowest 3\,s ADE and the highest lateral macro-F1 (Tab.~\ref{tab:indi2go_surround_ablation}). Adding all six views increases ADE from 4.411 to 4.564\,m and reduces lateral macro-F1 from 0.561 to 0.523. The six-view configuration has the highest mean longitudinal macro-F1, at 0.550 compared with 0.539 for pose history alone. This difference of 0.011 is small relative to the standard deviations across seeds. The comparison applies to fusion with one frozen image descriptor per camera. Section~\ref{sec:supp_camera_coverage} describes the architecture and qualitative examples. These values were obtained with the initial-release timestamps, under which the six image descriptors of a window are not simultaneous (Sec.~\ref{sec:supp_sync_audit}). {}

\subsection{Camera-Conditioned Motion-Prediction Architecture}
\label{sec:supp_camera_coverage}

In the six-camera ablation (Sec.~\ref{sec:supp_surround_ablation}),
image observations are represented using ImageNet-pretrained,
frozen ResNet-18 features~\cite{he2016resnet}. The selected camera
features are combined with the two-second scooter-state history using
a two-layer, four-head temporal Transformer~\cite{vaswani2017attention}.
The camera subsets are front, front/side, front/rear, and all six views,
with a pose-history-only reference. Only observations available at or
before the forecast anchor are included. All configurations predict
future planar positions and the lateral and longitudinal category targets.
Validation checkpoint selection is performed independently for seeds
17, 23, and 47. This Transformer evaluation is distinct from the
standalone MLP trajectory-forecasting experiment; their model families
and optimization objectives are specified in the respective main-paper
subsections. The quantitative comparisons {are in Tab.~\ref{tab:future_trajectory}
of the main paper and Tab.~\ref{tab:indi2go_surround_ablation}}.

\begin{figure*}[!t]
\centering
\includegraphics[width=0.88\textwidth]{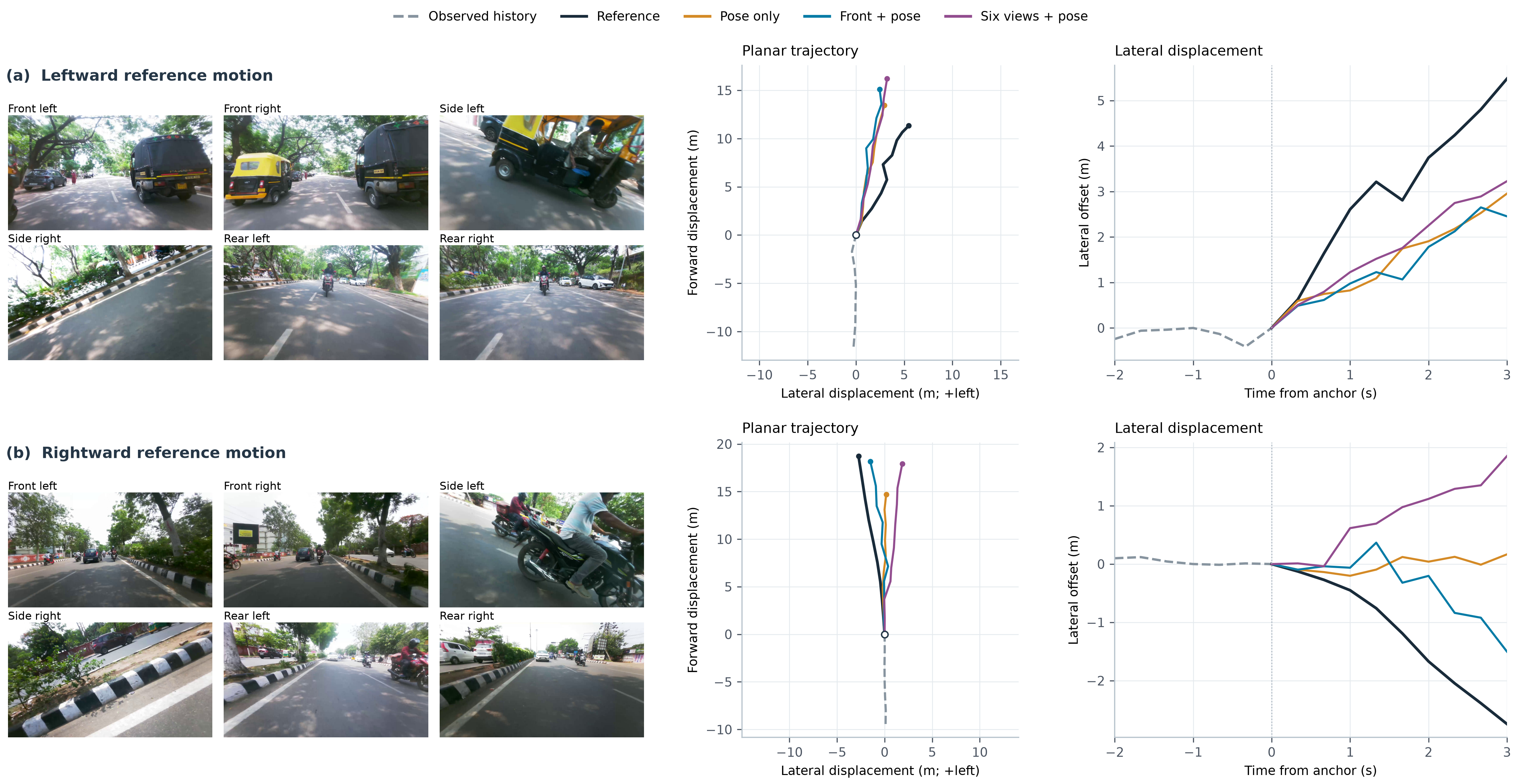}
\caption{Qualitative camera-conditioned prediction examples: six
surround-view images, ego-state history, and forecast and reference
trajectories. Camera observations precede or coincide with the anchor
used in Eq.~\eqref{eq:supp_anchor_relative}.}
\label{fig:supp_surround_forecasting_examples}
\end{figure*}

\subsection{Dynamic 4D Reconstruction}
\label{sec:dynamic_4d}

We evaluate NoPo4D~\cite{balice2026nopo4d} on temporal interpolation from the moving scooter. For a target keyframe $k$, the model receives 12 rectified RGB images, comprising all six views at $k-3$ and again at $k+3$. At the nominal 3\,Hz keyframe rate, these observations bracket the target by approximately one second on each side. The model reconstructs a dynamic Gaussian scene and renders six RGB images at the time of $k$, one per camera. The six recorded images at $k$ are withheld from the model and serve as scoring references.

The original NoPo4D formulation assumes a static camera rig~\cite{balice2026nopo4d}. Because the scooter cameras move through the world, we disable temporal pose averaging and interpolate each camera's predicted camera-to-world pose and intrinsics between the two input times. The renderer uses those interpolated parameters at the target time. Inference uses the rectified RGB inputs and model-predicted camera parameters.

We compare the public pretrained checkpoint with an Indi2Go fine-tuned checkpoint using the same rendering protocol. Fine-tuning updates only the Gaussian head and motion encoder with a Smooth L1 loss between rendered and recorded RGB images, using one window from each of the 756 training scenes. We select the checkpoint by mean PSNR on one window from each of the 17 validation recordings. Both checkpoints are frozen for evaluation on the same 162 test windows from 14 held-out recordings. We score 972 rendered images using PSNR, SSIM~\cite{wang2004ssim}, and AlexNet LPIPS~\cite{zhang2018lpips}. For each metric, we average the six target images within each scene, then average scenes within each recording and weight the recordings equally.

\begin{table}[H]
\centering
\caption{NoPo4D temporal interpolation on 972 target images from 162 test scenes and 14 recordings. Higher PSNR and SSIM and lower LPIPS indicate better agreement under the respective metrics.}
\label{tab:dynamic_4d}
\footnotesize
\setlength{\tabcolsep}{5pt}
\begin{tabular}{lrrr}
\toprule
Setting & PSNR (dB) & SSIM & LPIPS \\
\midrule
Native temporal & 12.844 & 0.5309 & \textbf{0.4927} \\
Fine-tuned & \textbf{15.0833} & \textbf{0.5687} & 0.5266 \\
\bottomrule
\end{tabular}
\end{table}

\subsection{Gravity-Aligned Road-Plane Depth Reference}
\label{sec:supp_depth_reference}

The zero-shot depth benchmark evaluates UniDepth V2~\cite{piccinelli2026unidepthv2},
  Depth Pro~\cite{bochkovskiy2025depthpro}, MoGe-2~\cite{wang2025moge2},
  Metric3D V2~\cite{hu2024metric3dv2}, and Depth Anything 3~\cite{lin2026depthanything3}.
  Mask2Former~\cite{cheng2022mask2former} supplies road masks, eroded by
  15 pixels and sampled every six pixels. For a calibrated camera center
  $\mathbf c$, ray $\mathbf r$, road-plane point $\mathbf p_0$, and
  plane normal $\mathbf n$ aligned with gravity, intersection with the
  plane is computed as
  \begin{equation}
  \lambda=\frac{\mathbf n^\mathsf T(\mathbf p_0-\mathbf c)}
  {\mathbf n^\mathsf T\mathbf r},\qquad
  \mathbf X=\mathbf c+\lambda\mathbf r.
  \label{eq:supp_roadplane_reference}
  \end{equation}
  Valid evaluation rays are at least $3^\circ$ below the horizon and
  produce reference depths between 1 and 8\,m. For a reference depth
  $d_j>0$ and predicted depth $\widehat d_j>0$ at a valid pixel, the
  threshold accuracy is
  \begin{equation}
  \delta_1=\frac{1}{N}\sum_{j=1}^N
  \mathbb{1}\!\left[\max\!\left(
  \frac{\widehat d_j}{d_j},\frac{d_j}{\widehat d_j}
  \right)<1.25\right].
  \label{eq:supp_depth_delta}
  \end{equation}
  Aligned estimates apply a separate per-image median scale, whereas
  native predictions retain their metric scale. The complete measured
  comparisons appear only in Tab.~\ref{tab:depth_main}; the original
  qualitative depth visualization is Fig.~\ref{fig:depth_qual}. Table~\ref{tab:depth_corrected} re-scores the cached predictions against a
  reference built with the ego attitude at each image's corrected capture time
  (Sec.~\ref{sec:supp_sync_audit}), which differs from the original attitude by a
  median of $1.2^\circ$ in pitch and roll (90th percentile $4.3^\circ$). The primary
  re-score covers approximately $1.4\times10^8$ road-reference points in the 5,265
  of the 7,269 images whose clock offset is confident and whose pose lies in a valid
  segment. The best model for each metric is unchanged: UniDepth V2 has the lowest
  aligned AbsRel (0.069), the highest aligned $\delta_1$ (94.8\%), and the lowest
  native AbsRel (0.166), and Depth Pro has the lowest aligned RMSE (0.448\,m). The
  same holds without the image filter (6,047 images). The ``released attitude,
  same points'' rows separate the effect of the smaller point set from that of the
  corrected reference.

  \input{tables/tab_depth_corrected.tex}

\subsection{Projected 3D-Detector Evaluation}
  \label{sec:supp_projected_detection}

  The transfer study uses pretrained DETR3D~\cite{wang2021detr3d},
  StreamPETR~\cite{wang2023streampetr}, Sparse4D v3~\cite{lin2023sparse4dv3},
  and BEVDepth~\cite{li2023bevdepth}. For each predicted 3D box,
  let $\mathbf X_k^E$ be its eight corners in the scooter ego frame.
  They are mapped to the rectified camera $C_i$ and projected by
  \begin{equation}
  \widetilde{\mathbf u}_{ik}\sim
  \mathbf K_i\,\mathbf P_3
  (\mathbf T^E_{C_i})^{-1}\widetilde{\mathbf X}_k^E,
  \label{eq:supp_box_projection}
  \end{equation}
  where $\mathbf P_3=[\mathbf I_3\mid\mathbf 0]$ selects Euclidean
  camera coordinates. A 2D rectangle enclosing projected corners is
  compared with the corresponding image annotation. The matched camera
  calibration and test recordings are those used in the projected-2D AP
  benchmark in Sec.~\ref{sec:det3d_protocol}; its numerical results
  remain in Tab.~\ref{tab:det3d_proj} of the main paper.
 The scenes of Tab.~\ref{tab:det3d_proj} of the main paper cannot be re-run
  with corrected groups, because only one of them has confident clock offsets for all six
  cameras. Tab.~\ref{tab:supp_det3d_corrected} therefore uses the human-verified test split
  of the 2D benchmark (11,508 labelled images, 44 scenes), restricted to the 6,313 labelled
  images (1,204 keyframes, 29 scenes) that have a valid corrected time and pose and an
  image from each other camera within 0.5\,s. The same images are run once with the
  original groups and once with the corrected groups (Sec.~\ref{sec:supp_sync_audit}). The scenes differ from
  those of Tab.~\ref{tab:det3d_proj}, so absolute scores differ from it (DETR3D AP50 13.3
  vs.\ 14.8) while the ranking is the same, and re-running DETR3D in the rebuilt
  environment reproduces its main-paper score (AP$_{50}$ 14.9 vs.\ 14.8).
  Correcting the groups raises AP50 and mAP4 of all four detectors and keeps the AP50
  ranking of the main paper (DETR3D, Sparse4D v3, StreamPETR, BEVDepth). Paired
  scene-bootstrap 95\% intervals exclude zero for DETR3D (AP50 $+8.3$, 6.1 to 11.0;
  mAP4 $+7.3$, 5.3 to 10.3), Sparse4D v3 (AP50 $+2.2$, 1.2 to 3.4; mAP4 $+2.2$, 1.6 to
  2.8) and the mAP4 of StreamPETR ($+1.0$, 0.6 to 1.4); the other changes are within
  noise. On the 4,821 time-consistent images (six images within 100\,ms) the gains are
  the same size (DETR3D AP50 13.3 to 21.8, Sparse4D v3 4.1 to 6.4). Cars keep the highest
  class AP50 for all four detectors with corrected groups, as in the main paper; with the
  original groups the near-zero BEVDepth scores are higher for heavy vehicles (1.6) than
  for cars (1.3). The scores measure image-plane agreement under our input adaptation
  (StreamPETR produces implausibly large boxes, and BEVDepth uses a PyTorch voxel
  pooling) and do not isolate a platform or domain effect.

  \begin{figure*}[!t]
  \centering
  \includegraphics[width=0.80\textwidth]{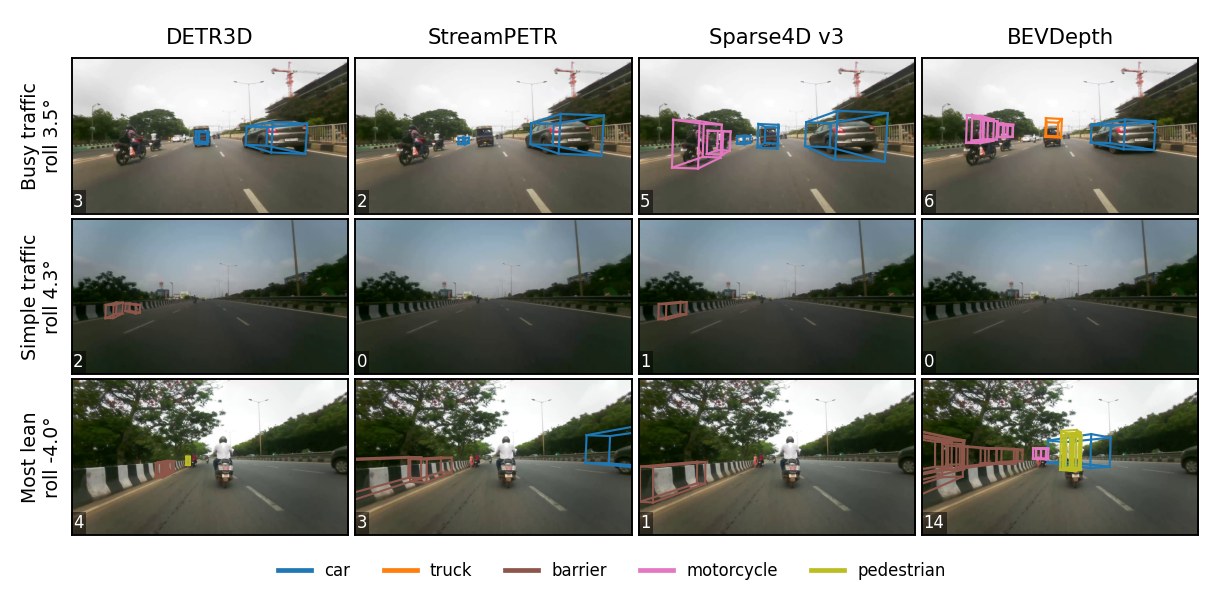}
  \caption{Qualitative projected 3D detections in the calibrated
  surround-view images}
  \label{fig:supp_det3d_examples}
  \end{figure*}

  \subsection{Zero-Shot Future-Trajectory Transfer}
  \label{sec:supp_uniad_transfer}

  The UniAD~\added{\cite{hu_uniad_2023}} transfer study follows
  an input-adaptation protocol distinct from in-domain 3\,Hz forecasting.
  The six Indi2Go channels are mapped deterministically to the pretrained
  camera slots, with a fixed navigation command and the synthetic LiDAR
  reference transform required by the inference interface. The target
  future is sampled at $+0.5,+1.0,\ldots,+3.0\,$s in the experiment's
  motion-aligned local frame. Evaluation uses anchors with the complete
  future horizon and the released timestamped ego-motion reference.
  Scores are reported in Tab.~\ref{tab:zero_shot_main} of the main paper. It was first run once on the original keyframe groups; scored as originally,
  it gives an ADE@3s of 8.23\,m on 50,239 anchors. Tab.~\ref{tab:zero_shot_main} of the main
  paper scores the same run with the reference anchored at the corrected capture time of the
  front-left image, which leaves 34,450 anchors (ADE@3s 8.83\,m, stand-still 14.81\,m).
  Table~\ref{tab:uniad_corrected} adds a re-run of UniAD on the corrected nearest-in-time
  groups (Sec.~\ref{sec:supp_sync_audit}), scored on the 35,068 eligible anchors of this run,
  which include the 34,450 anchors of the main paper. Correcting the
  groups lowers ADE@3s from 8.80\,m to 8.36\,m (paired difference $-0.44$\,m, 95\% CI
  $-0.53$ to $-0.35$) and FDE@3s from 14.97\,m to 14.28\,m; the heading error is
  unchanged. On the 18,738 time-consistent anchors (six images within 100\,ms) ADE@3s
  goes from 8.84\,m to 8.40\,m. UniAD remains far below the stand-still reference and
  above a constant-velocity reference (ADE@3s 5.24\,m on the eligible anchors, 5.04\,m on the time-consistent ones), so the conclusion of the main paper
  is
  unchanged and we report this as a transfer diagnostic, not a competitive benchmark.
  The result holds without the recording with flagged channels (8.52\,m), without
  anchors with a flagged pose (7.89\,m), and without anchors touching an invalid pose
  segment (8.51\,m). Re-running 102 scenes with the earlier pose stream changes
  ADE@3s by only 0.010\,m (3,980 anchors).

  \begin{table*}[!t]
  \centering
  \footnotesize
  \caption{Zero-shot UniAD ego trajectory: original groups vs.\ corrected groups
  (supplement). Except for the first row, ground truth is anchored at corrected image times; ADE/FDE in m. The 35,068
  anchors include the 34,450 of the main paper; ``within 100\,ms'' keeps anchors whose six
  images are within 100\,ms.}
  \label{tab:uniad_corrected}
  \setlength{\tabcolsep}{5pt}
  \begin{tabular}{lccccccc}
  \toprule
  Input groups & $n$ & ADE@1s & ADE@2s & ADE@3s & FDE@3s & Head.@3s ($^\circ$) & Stand-still ADE@3s \\
  \midrule
  Original, paper scoring, all anchors & 50239 & 3.54 & 5.89 & 8.23 & 14.08 & 23.90 & 15.26 \\
  Original, main-paper anchors (Tab.~\ref{tab:zero_shot_main}) & 34450 & 3.84 & 6.35 & 8.83 & 15.03 & 24.06 & 14.81 \\
  Original, all anchors & 35068 & 3.83 & 6.32 & 8.80 & 14.97 & 24.17 & 14.81 \\
  Corrected, all anchors (re-run) & 35068 & 3.60 & 5.99 & 8.36 & 14.28 & 24.17 & 14.81 \\
  \midrule
  Original, within 100\,ms & 18738 & 3.85 & 6.36 & 8.84 & 15.05 & 24.07 & 14.27 \\
  Corrected, within 100\,ms (re-run) & 18738 & 3.63 & 6.02 & 8.40 & 14.38 & 24.07 & 14.27 \\
  \bottomrule
  \end{tabular}
  \end{table*}

  \begin{table}[H]
  \centering
  \footnotesize
  \caption{Projected-2D AP (\%) of the pretrained 3D detectors on the same 6,313
  labelled images (1,204 keyframes, 29 scenes): original keyframe groups vs.\ corrected nearest-in-time
  groups. Rec50 is recall at IoU 0.5; mAP4 averages AP50 over Car, two-wheeler, heavy
  vehicle and pedestrian.}
  \label{tab:supp_det3d_corrected}
  \setlength{\tabcolsep}{4pt}
  \begin{tabular}{llcccc}
  \toprule
  Input groups & Detector & AP50 & AP10 & Rec50 & mAP4 \\
  \midrule
  Original groups & DETR3D      & 13.3 & 44.7 & 32.3 & 9.6 \\
   & StreamPETR  &  2.8 & 16.5 & 26.1 & 1.9 \\
   & Sparse4D v3 &  4.2 & 33.6 & 18.8 & 3.3 \\
   & BEVDepth    &  0.5 & 19.5 & 10.3 & 0.7 \\
  \midrule
  Corrected groups & DETR3D      & 21.6 & 52.4 & 40.3 & 16.9 \\
   & StreamPETR  &  3.7 & 16.4 & 28.2 & 2.9 \\
   & Sparse4D v3 &  6.4 & 39.7 & 24.3 & 5.5 \\
   & BEVDepth    &  0.6 & 22.4 & 11.4 & 1.0 \\
  \bottomrule
  \end{tabular}
  \end{table}

\section{2D Annotation Format and Linking}
\label{sec:supp_annotation_schema_notes}

The separately curated camera-local 2D object annotations follow
the COCO image/box conventions~\cite{lin2014coco}. The annotated
six-view mosaics are related to the released camera images by the
recorded mosaic tile geometry and timestamp-to-camera correspondence.
Each image-level box is associated with a camera-frame identity and
category within Indi2Go's object vocabulary. This task-specific
annotation representation is distinct from the optional 3D-style
annotation tables of the format, which are not populated in this release(Fig.~\ref{fig:dataset_schema}).

\begingroup
\normalcolor\color{black}
\section{2D Object Detection}
\label{sec:supp_det2d}

The 2D detection benchmark in Sec.~\ref{sec:det2d} and
Table~\ref{tab:main_v21} uses 200 scenes with human-verified
camera-level bounding boxes. This section documents the recording-disjoint
evaluation protocol and reports detector, class-wise, camera-wise, data-scaling,
and traffic-condition results on the same held-out partition. Pseudo-labels
for the other 880 scenes are kept separate from the reference annotations
used for evaluation.

\subsection{Dataset and Evaluation Protocol}
\label{sec:supp_det2d_data}

\paragraph{Human-verified reference annotations.}
The detection benchmark is constructed from 200 scenes with camera-local,
human-verified 2D bounding boxes. Annotation is performed on
six-view keyframe mosaics and transferred to individual rectified camera
coordinates; the evaluated boxes correspond to the original camera images,
not to boxes on a stitched panoramic image. Human review includes manually
traced or refined boxes for densely overlapping road users, especially the
prevalent motorized two-wheelers. In contrast to datasets that merge these
appearances into a single class, we preserve separate
\texttt{vehicle.motorcycle} and \texttt{vehicle.scooter} labels and retain
\texttt{vehicle.autorickshaw} as an additional traffic-specific category.
Small instances and partly occluded objects can require drawing missing
boxes and revising object boundaries as well as class identities. These
human-verified annotations, rather than model-generated labels from the
remaining scenes, are the ground truth for every result reported below.

\paragraph{Data partition.}
The benchmark comprises rectified $1920\times1080$ images from six
scooter-mounted cameras, with 38 road-user and roadside-object classes.
Scenes are partitioned at the recording level, yielding 125 training,
31 validation, and 44 test scenes (Table~\ref{tab:supp_v200_split}).
The 44 test scenes come from five recordings that are disjoint from
training and validation. There are 29 classes with nonzero test-set
reference support; AP is undefined for the other nine classes, rather
than being assigned zero. Class identities are mapped by their names,
since the category identifiers of the benchmark export are not assumed
to coincide with other detection vocabularies.

\begin{table}[H]
\centering
\caption{Recording-disjoint partition of the 200-scene 2D detection
benchmark. Images and bounding boxes are counted in camera coordinates.}
\label{tab:supp_v200_split}
\small
\setlength{\tabcolsep}{5pt}
\begin{tabular}{lrrr}
\toprule
Split & Scenes & Camera images & Boxes \\
\midrule
Train & 125 & 39,750 & 75,617 \\
Validation & 31 & 10,404 & 18,342 \\
Test & 44 & 11,508 & 27,288 \\
\midrule
Total & 200 & 61,662 & 121,247 \\
\bottomrule
\end{tabular}
\end{table}

\paragraph{Input geometry and checkpoint selection.}
All detectors operate on an individual camera view; no panoramic
mosaic is used as a detector input. Native $16{:}9$ geometry is retained
through aspect-ratio-preserving resizing and padding. For the
transformer configurations evaluated at the larger rectangular scale,
$1920\times1080$ images are resized to $1280\times720$ and padded to
$1280\times736$. YOLO26-M uses the corresponding 1280-pixel rectangular
input. The larger RF-DETR-L and DEIMv2-X configurations were selected
using validation data; the input scales of the medium configurations
follow the main-paper protocol and were not retuned on the test set.
The separate YOLO26-L experiment in Table~\ref{tab:main_v21} uses a
960-pixel input and three training epochs.

Fine-tuning starts from the respective COCO-pretrained weights, replacing
the detection head with the 38-class benchmark vocabulary. For each of
the three primary training seeds (0--2), the checkpoint is selected using
validation AP$_{50:95}$ only and evaluated once on the held-out test
partition. The common COCO evaluator receives boxes mapped back to the
original $1920\times1080$ coordinates, retaining at most 100 detections
per image. We report AP$_{50:95}$, AP$_{50}$, and AP$_{75}$ in percentage
points. Mean $\pm$ sample standard deviation (SD; divisor $n-1$) is
reported for the three-seed experiments; SD measures training-seed
variation and is not a confidence interval. The seven-class COCO-overlap
subset, Z7, comprises bicycle, car, motorcycle, bus, truck, cow, and dog;
the more general \emph{person} category is not included in this mapping.

\paragraph{Training implementation.}
The principal three-seed models follow their respective COCO-pretrained
training recipes with a 38-class output vocabulary. The large RF-DETR-L
and DEIMv2-X configurations use a maximum of 12 epochs and
validation-based early stopping (patience three), with exponential
moving-average checkpoints selected by the best validation
AP$_{50:95}$. RF-DETR-L uses a batch of 16, a $10^{-4}$ learning
rate, and random horizontal flips. DEIMv2-X uses a total batch of 32,
a $5\times10^{-4}$ base learning rate, a $10^{-5}$ backbone learning
rate, and photometric augmentation; mosaic, mixup, copy-blend, and
random cropping are not used in this configuration. Medium-model
configurations retain their documented model-family recipes. The
reported standard deviations describe seed sensitivity and should
not be read as confidence intervals.

\subsection{Detector Comparisons}
\label{sec:supp_det2d_architecture}

The main-paper comparison includes YOLO26-M, RF-DETR-M/L, D-FINE-M,
and DEIMv2-X, together with a separately trained YOLO26-L model
(Table~\ref{tab:main_v21}). As an additional capacity comparison
within the DEIMv2 family, we evaluate DEIMv2-M with the same held-out
reference set and three-seed aggregation. Its full-vocabulary AP is
$19.52\pm0.65$, compared with $21.23\pm0.73$ for DEIMv2-X
(Table~\ref{tab:supp_v200_deim}). This $1.71$-point difference is
observational: model capacity, initialization, and optimization
configurations are not isolated by a compute-matched ablation.

\begin{table*}[!t]
\centering
\caption{Fine-tuned 2D detector performance on the human-verified 44-scene test split. All results are COCO AP points. Values are mean $\pm$ sample SD over seeds 0--2 ($n=3$). Z7 comprises seven COCO-overlap classes. The final column gives the fine-tuning gain over the corresponding COCO-pretrained zero-shot checkpoint.}
\label{tab:supp_v200_deim}
\footnotesize
\setlength{\tabcolsep}{4pt}
\renewcommand{\arraystretch}{1.08}
\begin{tabular*}{0.92\textwidth}{@{\extracolsep{\fill}}lrrrrrr@{}}
\toprule
Model & Test AP & AP$_{50}$ & AP$_{75}$ & Z7 AP & Zero-shot Z7 & Z7 gain \\
\midrule
DEIMv2-X & $21.23\pm0.73$ & $26.84\pm1.03$ & $23.49\pm1.02$ & $47.50\pm0.34$ & 33.57 & $13.93\pm0.34$ \\
RF-DETR-L & $19.44\pm0.79$ & $24.78\pm0.98$ & $21.27\pm1.02$ & $48.00\pm0.44$ & 36.14 & $11.86\pm0.44$ \\
RF-DETR-M & $20.31\pm0.22$ & $25.91\pm0.34$ & $22.47\pm0.08$ & $48.36\pm0.29$ & 34.04 & $14.32\pm0.29$ \\
DEIMv2-M & $19.52\pm0.65$ & $24.18\pm0.94$ & $21.95\pm0.69$ & $46.91\pm0.44$ & 32.43 & $14.48\pm0.44$ \\
D-FINE-M & $18.75\pm0.70$ & $23.12\pm0.99$ & $20.93\pm0.75$ & $46.00\pm0.38$ & 27.94 & $18.06\pm0.38$ \\
YOLO26-M & $17.33\pm0.55$ & $21.68\pm0.67$ & $19.24\pm0.57$ & $42.98\pm1.41$ & 31.04 & $11.94\pm1.41$ \\
\bottomrule
\end{tabular*}
\end{table*}

The fine-tuned Z7 APs are $48.36\pm0.29$ for RF-DETR-M,
$48.00\pm0.44$ for RF-DETR-L, $47.50\pm0.34$ for DEIMv2-X,
$46.91\pm0.44$ for DEIMv2-M, $46.00\pm0.38$ for D-FINE-M,
and $42.98\pm1.41$ for YOLO26-M. Their corresponding COCO-pretrained
Z7 baselines are 34.04, 36.14, 33.57, 32.43, 27.94, and 31.04,
respectively.

On Z7, the main-paper models improve by $11.86$--$18.06$ AP points
over their corresponding frozen COCO-pretrained baselines, with the
largest gain for D-FINE-M. The gains quantify the paired evaluation
under the benchmark's category mapping; they should not be interpreted
as the effect of additional training data alone. Two further completed
seeds for DEIMv2-X and RF-DETR-L give, respectively,
$21.56\pm0.76$ and $19.58\pm0.88$ AP ($n=5$). These additional runs
are a stability check; the main-paper comparison consistently uses
the first three seeds.

\paragraph{Open-vocabulary transfer.}
The open-vocabulary baselines use class-name prompts and no Indi2Go
fine-tuning. As detailed in Table~\ref{tab:supp_v200_ov}, OWLv2 yields
the highest full-vocabulary AP among the three evaluated models (14.41),
whereas Grounding DINO Base is slightly higher on Z7 (29.88 versus
29.53). The change in ranking is consistent with the different class
support of the two metrics; it does not demonstrate a statistically
resolved difference between the methods.

\begin{table}[H]
\centering
\caption{Open-vocabulary zero-shot detection on the 44-scene test
partition. Each model is evaluated once; scores are COCO AP points.}
\label{tab:supp_v200_ov}
\small
\setlength{\tabcolsep}{4pt}
\begin{tabular}{lrr}
\toprule
Model & Full AP ($K=29$) & Z7 AP \\
\midrule
Grounding DINO Tiny & 4.47 & 13.74 \\
Grounding DINO Base & 11.15 & \textbf{29.88} \\
OWLv2 & \textbf{14.41} & 29.53 \\
\bottomrule
\end{tabular}
\end{table}

\subsection{Qualitative Detection Results and Privacy Protection}
\label{sec:supp_det2d_qualitative}

\paragraph{Qualitative detection comparison.}
Figure~\ref{fig:supp_det2d_qualitative} compares the human-verified CVAT
reference boxes with predictions from the COCO-pretrained YOLO26-L
checkpoint and the separately fine-tuned three-epoch YOLO26-L checkpoint.
Each row presents the same front-left camera image across all three
columns. The comparison illustrates the challenges of fine-grained
road-user recognition in heterogeneous Indian traffic. In the first
example, the reference distinguishes scooters from adult pedestrians,
whereas the zero-shot detector uses the broader COCO categories and the
fine-tuned detector still misclassifies a scooter as a bicycle. In the
second example, the fine-tuned detector localizes two motorcycles but
misses the central motorcycle and the nearby pedestrian. These examples
highlight residual classification and localization failures under
viewpoint variation, motion blur, and crowded roadside conditions.
The displayed predictions are not manually corrected or replaced with
reference annotations. The examples are illustrative and do not constitute
an additional quantitative evaluation or establish a uniform improvement
from fine-tuning.

\begin{figure*}[!t]
    \centering
    \includegraphics[width=\textwidth]{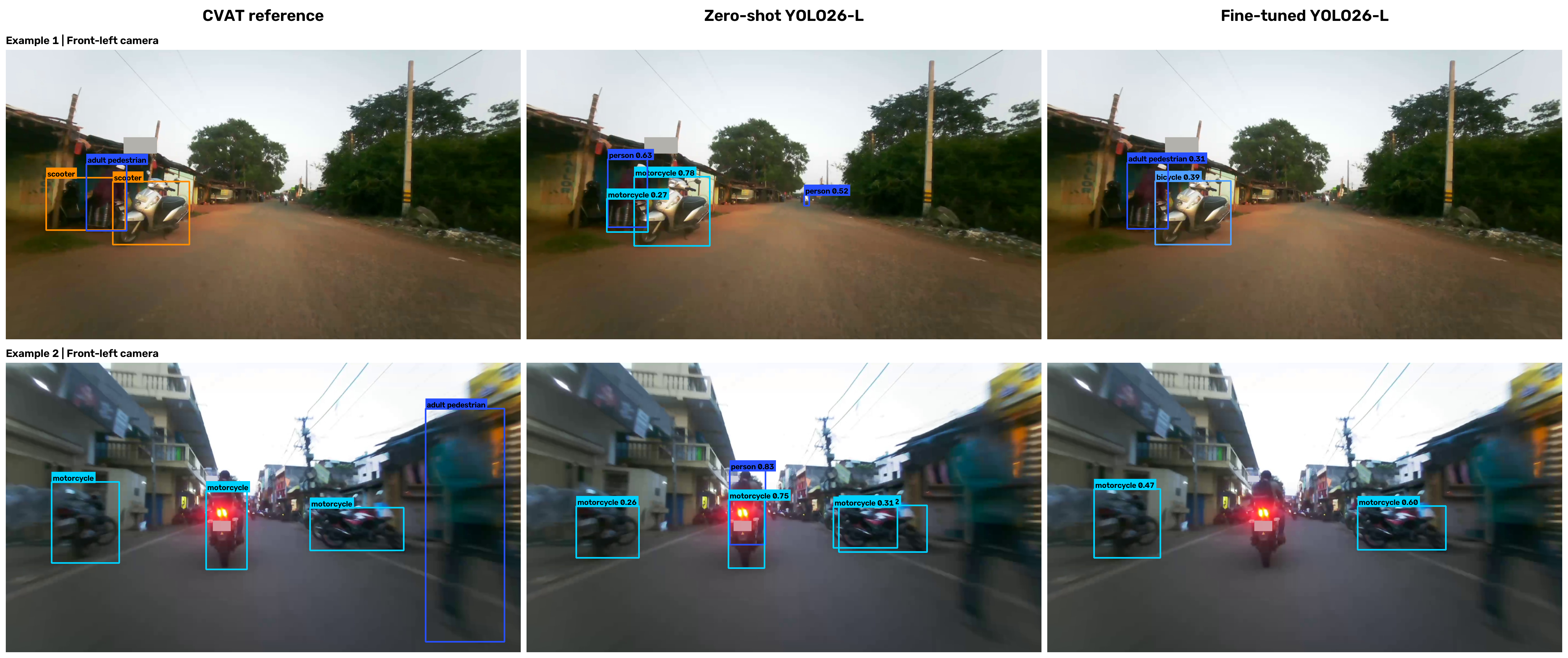}
    \caption{Qualitative 2D detection comparison on two front-left
    camera images from Indi2Go. Left: human-verified CVAT reference
    boxes. Middle: COCO-pretrained zero-shot YOLO26-L predictions.
    Right: fine-tuned YOLO26-L predictions using the Indi2Go vocabulary.
    Prediction labels retain their detector confidence scores. The
    examples illustrate fine-grained category confusion and missed
    detections in unstructured traffic. Identical localized privacy
    masks are used across the three columns for visualization.}
    \label{fig:supp_det2d_qualitative}
\end{figure*}

\paragraph{Human-assisted fine-grained anonymization.}
Reliable anonymization is particularly challenging for scooter-mounted
cameras in dense and unstructured Indian traffic. In our current
workflow, existing automatic face and license-plate anonymizers do not
consistently identify all sensitive regions. Small or distant faces,
partially occluded road users, oblique or non-standard license plates,
and motion blur can lead to missed detections or imprecise privacy masks.
To improve coverage, we combine automatic detection proposals at multiple
image scales with localized redaction and manual review. Human annotators
are involved in inspecting candidate masks, identifying missed sensitive
regions, and refining mask boundaries while preserving the surrounding
road-user appearance and scene context as far as possible.
The redacted regions visible in Fig.~\ref{fig:supp_det2d_qualitative}
illustrate the privacy treatment used for the qualitative visualization.
This privacy review is distinct from the human verification of 2D
object-detection labels. Fine-grained anonymization and verification of
the remaining images are ongoing, and the visual examples should not
be interpreted as evidence of complete dataset-wide anonymization or as
a quantitative comparison of anonymization methods. Final public release
of the imagery is subject to completion of the privacy review.

\subsection{Camera-Wise Detection Performance}
\label{sec:supp_v200_percamera}

We next disaggregate the test predictions by camera without changing
model checkpoints or evaluation settings. Each of the six cameras
contributes 1,918 test images. Table~\ref{tab:supp_v200_percamera}
reports AP$_{50:95}$ separately for each view. For DEIMv2-X,
back-right reaches 34.87 AP, while front-left reaches 26.12 AP;
RF-DETR-M attains its highest camera-specific AP on side-left (31.32).
These measurements expose substantial view dependence, but do not
measure an isolated effect of camera orientation: scene content,
object size, and category support differ across views. In particular,
COCO evaluation excludes classes with no reference objects in a given
camera. Hence the camera-wise figures average over different class
sets and are not directly comparable to the overall 29-class AP.

\begin{table*}[!t]
\centering
\caption{Camera-specific test AP$_{50:95}$ (three-seed mean). Each camera contributes 1,918 test images. Absent classes are excluded independently within each view; these values are not directly comparable to the overall 29-class test AP.}
\label{tab:supp_v200_percamera}
\footnotesize
\setlength{\tabcolsep}{5pt}
\renewcommand{\arraystretch}{1.10}
\begin{tabular*}{0.92\textwidth}{@{\extracolsep{\fill}}lrrrrrr@{}}
\toprule
Model & Back-left & Back-right & Front-left & Front-right & Side-left & Side-right \\
\midrule
DEIMv2-X & 32.65 & 34.87 & 26.12 & 28.11 & 30.82 & 29.15 \\
RF-DETR-L & 27.15 & 27.95 & 25.35 & 27.43 & 26.39 & 25.95 \\
RF-DETR-M & 27.46 & 29.28 & 26.31 & 29.65 & 31.32 & 27.85 \\
DEIMv2-M & 26.30 & 28.19 & 25.54 & 27.84 & 25.84 & 26.53 \\
D-FINE-M & 28.76 & 28.28 & 26.14 & 28.74 & 23.77 & 25.91 \\
YOLO26-M & 23.94 & 25.13 & 22.93 & 24.87 & 24.97 & 22.77 \\
\bottomrule
\end{tabular*}
\end{table*}

\subsection{Class-Wise Detection Performance}
\label{sec:supp_v200_perclass}

Table~\ref{tab:supp_v200_perclass} reports representative class-level
scores covering common vehicles, two-wheelers, pedestrians, and
lower-support road users. Cars, motorcycles, and autorickshaws exhibit
substantially higher AP than several visually ambiguous or less
frequent classes. For example, DEIMv2-X obtains 80.31 AP on cars,
71.97 on motorcycles, and 48.16 on scooters, but 8.47 on dogs.
This variation is masked by the aggregate AP; it also reflects
class-specific test support and should not be interpreted as a
controlled comparison of object difficulty. The 29-class summary
includes all classes with test reference boxes, not only those shown
here.

\begin{table}[H]
\centering
\caption{Selected class-specific test AP (mean across seeds 0--2).
These values are individual-class diagnostics, not a second aggregate
or a different class vocabulary.}
\label{tab:supp_v200_perclass}
\scriptsize
\setlength{\tabcolsep}{3.5pt}
\begin{tabular}{lrrr}
\toprule
Class & DEIMv2-X & RF-DETR-M & YOLO26-M \\
\midrule
Adult pedestrian & 46.81 & 44.76 & 45.15 \\
Bicycle & 67.89 & 65.20 & 63.69 \\
Motorcycle & 71.97 & 70.15 & 71.13 \\
Scooter & 48.16 & 48.24 & 45.95 \\
Autorickshaw & 69.22 & 68.73 & 69.90 \\
Car & 80.31 & 79.04 & 79.22 \\
Truck & 27.73 & 28.49 & 26.30 \\
Dog & 8.47 & 17.02 & 8.90 \\
\bottomrule
\end{tabular}
\end{table}

\subsection{Training Data Scaling}
\label{sec:supp_v200_scaling}

To examine training-data sensitivity, D-FINE-M is retrained on nested,
scene-level subsets of the 125 training scenes while keeping the
validation and test partitions fixed. The experiment retains the
model-family training recipe and the same 38-class evaluation.
Table~\ref{tab:supp_v200_scaling} shows that mean test AP rises from
$12.44\pm0.05$ with 10\% of the training subset to $18.75\pm0.70$
with all training scenes, a difference of 6.31 points. The marginal
gain between 75\% and 100\% is 0.16 points, smaller than the observed
training-seed variation. Because the subsets have different numbers
of training images and optimization steps, the curve is not
compute-matched and does not establish a performance ceiling.

\begin{table}[H]
\centering
\caption{D-FINE-M performance with nested subsets of the 125 training
scenes. The same held-out test split is used throughout. Values are
test AP$_{50:95}$, mean $\pm$ sample SD over three seeds.}
\label{tab:supp_v200_scaling}
\small
\setlength{\tabcolsep}{4pt}
\begin{tabular}{rrr}
\toprule
Training fraction & Camera images & Test AP \\
\midrule
10\% & 4,230 & $12.44\pm0.05$ \\
25\% & 10,344 & $15.47\pm0.63$ \\
50\% & 19,464 & $16.73\pm0.55$ \\
75\% & 29,718 & $18.59\pm0.55$ \\
100\% & 39,750 & $18.75\pm0.70$ \\
\bottomrule
\end{tabular}
\end{table}

\subsection{Performance Across Traffic Conditions}
\label{sec:supp_v200_conditions}

We examine performance differences across time of day, camera group,
traffic density, two-wheeler prevalence, and object size. The strata
are defined from recording metadata or reference boxes, with
thresholds fixed before inspection of the test results. Table~\ref{tab:supp_v200_conditions_ap} reports the corresponding AP values.
Each subset is evaluated with the classes supported by its own
reference boxes; comparisons across differently supported subsets
are therefore descriptive. Images without any reference boxes
(2,884) do not enter the density and two-wheeler-prevalence strata.

\begin{table*}[!t]
\centering
\caption{Condition-stratified AP$_{50:95}$ (three-seed mean $\pm$ sample SD). Counts give camera images and reference boxes; class support varies across strata. Models are ordered as in the header.}
\label{tab:supp_v200_conditions_ap}
\footnotesize
\setlength{\tabcolsep}{3.5pt}
\renewcommand{\arraystretch}{1.06}
\begin{tabular*}{0.96\textwidth}{@{\extracolsep{\fill}}llrrrrrr@{}}
\toprule
Condition & Images/boxes & DEIM-X & RF-L & RF-M & DEIM-M & D-FINE & YOLO-M \\
\midrule
\textit{Time}: Morning & 8,760/25,636 & $22.0\pm0.5$ & $21.1\pm0.9$ & $21.8\pm0.7$ & $21.8\pm0.5$ & $20.6\pm1.0$ & $19.2\pm0.8$ \\
Late afternoon & 2,748/1,652 & $32.3\pm0.6$ & $29.7\pm0.7$ & $30.5\pm1.7$ & $28.7\pm0.5$ & $26.6\pm0.8$ & $25.4\pm0.0$ \\
\addlinespace[1.5pt]
\textit{Camera}: Front & 3,836/10,867 & $25.5\pm1.1$ & $23.9\pm1.4$ & $25.4\pm0.2$ & $25.0\pm1.7$ & $25.5\pm1.0$ & $21.9\pm1.4$ \\
Side & 3,836/4,571 & $22.2\pm1.0$ & $19.8\pm1.4$ & $22.0\pm0.6$ & $20.3\pm0.7$ & $18.4\pm1.5$ & $16.9\pm0.8$ \\
Back & 3,836/11,850 & $31.6\pm1.1$ & $26.4\pm0.8$ & $26.9\pm1.0$ & $26.1\pm0.1$ & $26.7\pm0.8$ & $23.5\pm0.7$ \\
\addlinespace[1.5pt]
\textit{Density}: Low & 3,728/5,511 & $20.6\pm2.5$ & $18.7\pm0.9$ & $19.7\pm0.5$ & $19.0\pm1.5$ & $17.5\pm1.1$ & $15.1\pm0.1$ \\
Mid & 1,672/5,016 & $32.4\pm2.0$ & $26.2\pm1.6$ & $29.6\pm1.6$ & $26.3\pm3.0$ & $26.4\pm2.8$ & $22.9\pm2.2$ \\
High & 3,224/16,761 & $30.4\pm1.0$ & $28.4\pm1.1$ & $30.3\pm1.0$ & $31.1\pm2.8$ & $27.5\pm0.8$ & $25.4\pm1.2$ \\
\addlinespace[1.5pt]
\textit{Two-wheeler}: None & 1,991/4,145 & $20.0\pm0.2$ & $17.8\pm1.9$ & $19.1\pm0.9$ & $16.9\pm0.5$ & $16.1\pm0.8$ & $14.7\pm0.9$ \\
Minority & 1,741/8,226 & $32.0\pm1.8$ & $26.6\pm0.7$ & $28.3\pm0.5$ & $26.3\pm1.7$ & $28.8\pm1.6$ & $23.6\pm0.9$ \\
Majority & 2,754/10,904 & $29.2\pm1.1$ & $27.1\pm0.2$ & $27.5\pm0.7$ & $27.9\pm0.5$ & $27.1\pm0.9$ & $24.3\pm0.6$ \\
All & 2,138/4,013 & $66.1\pm0.7$ & $65.3\pm0.3$ & $66.0\pm0.6$ & $64.3\pm0.7$ & $64.5\pm0.9$ & $63.4\pm1.4$ \\
\addlinespace[1.5pt]
\textit{Size}: Small & 3,966/6,430 & $21.6\pm0.6$ & $21.2\pm0.4$ & $21.4\pm0.3$ & $21.7\pm0.5$ & $20.9\pm0.7$ & $22.1\pm0.3$ \\
Medium & 4,874/8,715 & $27.3\pm1.4$ & $27.0\pm1.2$ & $26.9\pm0.3$ & $28.1\pm1.2$ & $25.8\pm1.2$ & $24.9\pm1.1$ \\
Large & 6,534/12,143 & $25.8\pm0.9$ & $23.6\pm1.3$ & $24.5\pm0.6$ & $22.7\pm1.1$ & $22.3\pm1.4$ & $19.7\pm0.6$ \\
\bottomrule
\end{tabular*}
\end{table*}

The late-afternoon subset comprises a single recording
(\texttt{28-05-26-16}, beginning at 18:22 local time), whereas the
morning subset spans four recordings. Thus, differences in their AP
cannot isolate illumination from route and recording identity.
Similarly, the ``all'' two-wheeler stratum contains only three
classes with test reference objects; its numerically high AP is
not directly comparable to more diverse strata. The size results
show that no single detector is uniformly best: YOLO26-M reaches
$22.1\pm0.3$ AP in the small-object stratum, while DEIMv2-M reaches
$28.1\pm1.2$ in the medium-object stratum. These are conditional
measurements, not significance-tested model rankings.

\paragraph{Truncation and overlap-derived occlusion.}
For object-level diagnostics, truncation is defined by a reference box
within 8 pixels of an image border, and occlusion is approximated by
overlap with lower image boxes. We report recall at 100 predictions
(AR@100) rather than AP on these box-restricted subsets: unmatched
predictions cannot be attributed to one object-level subset without
introducing an artificial false-positive accounting rule. The recall
figures in Table~\ref{tab:supp_v200_conditions_ar} are conditional on
class and box support. In particular, larger recall in the
heavy-overlap stratum does not establish that occlusion improves
performance.

\begin{table*}[!t]
\centering
\caption{Box-restricted recall AR@100 (three-seed mean $\pm$ sample SD) for truncation and overlap-derived occlusion. These figures are not whole-image AP.}
\label{tab:supp_v200_conditions_ar}
\footnotesize
\setlength{\tabcolsep}{3.5pt}
\renewcommand{\arraystretch}{1.06}
\begin{tabular*}{0.96\textwidth}{@{\extracolsep{\fill}}llrrrrrr@{}}
\toprule
Condition & Images/boxes & DEIM-X & RF-L & RF-M & DEIM-M & D-FINE & YOLO-M \\
\midrule
\textit{Border}: Not truncated & 7,728/21,860 & $60.1\pm0.9$ & $59.6\pm1.7$ & $61.8\pm1.1$ & $60.4\pm1.1$ & $57.0\pm0.9$ & $41.5\pm0.9$ \\
Truncated & 4,122/5,428 & $68.5\pm1.9$ & $62.9\pm2.2$ & $64.5\pm0.6$ & $65.6\pm1.1$ & $67.4\pm2.9$ & $48.5\pm3.7$ \\
\addlinespace[1.5pt]
\textit{Overlap}: Low & 8,624/25,382 & $60.9\pm0.6$ & $60.5\pm1.4$ & $62.2\pm0.7$ & $60.8\pm0.7$ & $58.3\pm0.9$ & $42.1\pm1.4$ \\
Partial & 1,041/1,169 & $67.8\pm0.9$ & $62.6\pm1.6$ & $67.0\pm1.4$ & $64.9\pm1.9$ & $60.6\pm1.4$ & $50.0\pm2.1$ \\
Heavy & 656/737 & $79.4\pm0.8$ & $77.7\pm0.5$ & $78.2\pm1.2$ & $76.7\pm0.9$ & $78.2\pm0.8$ & $73.8\pm2.6$ \\
\bottomrule
\end{tabular*}
\end{table*}

Taken together, the supplementary experiments expose substantial
variation across cameras, class categories, and operating conditions
under a fixed held-out benchmark. They complement the aggregate
results in Table~\ref{tab:main_v21}; none of the stratified AP
measurements substitutes for its common 29-class evaluation.

\subsection{Pseudo-Label Generation and Human Review}
\label{sec:supp_v200_pseudolabels}

The quantitative evaluation above uses only the human-verified 200-scene
benchmark. As described in Sec.~\ref{sec:det2d}, camera-local 2D
pseudo-labels for the remaining 880 scenes are generated using the
separately fine-tuned three-epoch YOLO26-L checkpoint
(\texttt{best.pt}), on individual rectified views with
aspect-ratio-preserving 960-pixel inputs. These proposals
are not counted as test ground truth or included in the benchmark AP scores.
AI-assisted annotation and subsequent human correction and verification of
these additional scenes are ongoing. Human annotators review class identities,
box geometry, false positives, and missed instances, with particular attention
to motorcycles, scooters, autorickshaws, and small or crowded road users.
Accordingly, the released label provenance distinguishes model proposals
from completed CVAT corrections rather than assigning a common
human-verification status to all 880 scenes.

\paragraph{Camera-level JSON representation.}
The camera-level 2D annotation format uses two files under
\texttt{v1.0-trainval/}: \texttt{instances\_2d.json} and
\texttt{annotations\_2d\_status.json}; the latter records scene-level
coverage and review status. The first file uses COCO-style
\texttt{images}, \texttt{annotations}, and \texttt{categories} arrays.
Each image record is linked to the released camera image through its
\texttt{sample\_data\_token} and camera channel; each bounding box is
represented as \texttt{[x,y,w,h]} in the individual rectified
$1920\times1080$ image. Pseudo-label boxes retain their detector
confidence, and source/review-status fields distinguish unreviewed predictions
from CVAT-corrected annotations. The frozen benchmark reference labels remain
in the separate \texttt{verified200\_human\_v1} COCO split files. 

\endgroup

%% file: tables/tab_depth_corrected.tex
\begin{table*}[!t]
\centering
\footnotesize
\caption{Zero-shot monocular depth: road-plane reference with released vs corrected timing of the ego attitude.}
\label{tab:depth_corrected}
\setlength{\tabcolsep}{3pt}
\begin{tabular}{llrrrrrrr}
\toprule
Method & Reference & A-AbsRel$\downarrow$ & A-RMSE$\downarrow$ & A-$\delta_1\uparrow$ & N-AbsRel$\downarrow$ & N-$\delta_1\uparrow$ & Images & Points \\
\midrule
UniDepth V2 & Released attitude & 0.074 & 0.587 & 94.1 & 0.173 & 66.5 & 7{,}269 & 199{,}620{,}903 \\ %
 & Corrected attitude & 0.069 & 0.541 & 94.8 & 0.166 & 68.9 & 5{,}265 & 142{,}119{,}750 \\ %
 & \quad released attitude, same points & 0.072 & 0.549 & 94.5 & 0.168 & 68.2 & 5{,}265 & 142{,}119{,}750 \\ %
 & Corrected, no image filter & 0.069 & 0.532 & 94.9 & 0.171 & 66.9 & 6{,}047 & 163{,}764{,}675 \\ %
 & \quad released attitude, same points & 0.071 & 0.541 & 94.6 & 0.172 & 66.6 & 6{,}047 & 163{,}764{,}675 \\ %
\midrule
Depth Pro & Released attitude & 0.078 & 0.489 & 93.4 & 0.220 & 60.0 & 7{,}269 & 199{,}620{,}903 \\ %
 & Corrected attitude & 0.072 & 0.448 & 94.5 & 0.210 & 62.0 & 5{,}265 & 142{,}119{,}750 \\ %
 & \quad released attitude, same points & 0.075 & 0.460 & 94.1 & 0.212 & 61.8 & 5{,}265 & 142{,}119{,}750 \\ %
 & Corrected, no image filter & 0.073 & 0.450 & 94.2 & 0.218 & 60.3 & 6{,}047 & 163{,}764{,}675 \\ %
 & \quad released attitude, same points & 0.076 & 0.462 & 93.9 & 0.219 & 60.2 & 6{,}047 & 163{,}764{,}675 \\ %
\midrule
MoGe-2 & Released attitude & 0.084 & 0.550 & 92.8 & 0.663 & 12.5 & 7{,}269 & 199{,}620{,}903 \\ %
 & Corrected attitude & 0.078 & 0.509 & 93.7 & 0.652 & 12.8 & 5{,}265 & 142{,}119{,}750 \\ %
 & \quad released attitude, same points & 0.081 & 0.519 & 93.4 & 0.654 & 13.1 & 5{,}265 & 142{,}119{,}750 \\ %
 & Corrected, no image filter & 0.079 & 0.504 & 93.6 & 0.656 & 12.5 & 6{,}047 & 163{,}764{,}675 \\ %
 & \quad released attitude, same points & 0.082 & 0.515 & 93.2 & 0.658 & 12.8 & 6{,}047 & 163{,}764{,}675 \\ %
\midrule
Metric3D V2 & Released attitude & 0.093 & 1.241 & 92.1 & 0.680 & 31.6 & 7{,}269 & 199{,}620{,}903 \\ %
 & Corrected attitude & 0.090 & 1.323 & 92.9 & 0.722 & 29.6 & 5{,}265 & 142{,}119{,}750 \\ %
 & \quad released attitude, same points & 0.094 & 1.324 & 92.6 & 0.725 & 29.8 & 5{,}265 & 142{,}119{,}750 \\ %
 & Corrected, no image filter & 0.090 & 1.250 & 92.7 & 0.690 & 31.7 & 6{,}047 & 163{,}764{,}675 \\ %
 & \quad released attitude, same points & 0.093 & 1.252 & 92.4 & 0.693 & 31.8 & 6{,}047 & 163{,}764{,}675 \\ %
\midrule
Depth Anything 3 & Released attitude & 0.091 & 0.516 & 92.0 & 0.359 & 35.7 & 7{,}269 & 199{,}620{,}903 \\ %
 & Corrected attitude & 0.086 & 0.478 & 93.1 & 0.357 & 35.6 & 5{,}265 & 142{,}119{,}750 \\ %
 & \quad released attitude, same points & 0.088 & 0.489 & 92.7 & 0.359 & 35.5 & 5{,}265 & 142{,}119{,}750 \\ %
 & Corrected, no image filter & 0.088 & 0.480 & 92.8 & 0.353 & 36.2 & 6{,}047 & 163{,}764{,}675 \\ %
 & \quad released attitude, same points & 0.090 & 0.491 & 92.4 & 0.355 & 36.0 & 6{,}047 & 163{,}764{,}675 \\ %
\bottomrule
\end{tabular}
\par\vspace{1pt}{\scriptsize
Cached predictions re-scored at the same pixels (models not re-run); reference recomputed with ego attitude at the corrected capture time; newly valid pixels not added.\par
Primary: timing/pose filter (fixed before results); no-filter rows: scorer as first frozen. A/N = median-aligned/native; RMSE in m.\par
}
\end{table*}